\documentclass{article} 
\usepackage{iclr2024_conference,times}

\usepackage{amsmath,amsfonts,bm}

\def\eqref#1{equation~\ref{#1}}

\def\1{\bm{1}}

\DeclareMathAlphabet{\mathsfit}{\encodingdefault}{\sfdefault}{m}{sl}
\SetMathAlphabet{\mathsfit}{bold}{\encodingdefault}{\sfdefault}{bx}{n}

\usepackage{hyperref}
\usepackage{url}
\usepackage[utf8]{inputenc} 
\usepackage[T1]{fontenc}    
\usepackage{hyperref}       
\usepackage{url}            
\usepackage{booktabs}       
\usepackage{amsfonts}       
\usepackage{nicefrac}       
\usepackage{microtype}      
\usepackage{xcolor}         
\usepackage{makecell}
\usepackage{graphicx}
\usepackage{amsmath}
\usepackage{multirow}
\usepackage{wrapfig}
\usepackage{algorithm}
\usepackage{algpseudocode}
\usepackage{enumitem}
\usepackage{wrapfig}
\usepackage{float}
\usepackage{graphicx}
\usepackage{caption}
\usepackage{color}
\usepackage{colortbl}

\def\ie{\emph{i.e., }}

\title{TransNormerLLM: A Faster and Better Large Language Model with Improved TransNormer}

\author{
{\normalsize
Zhen Qin$^\sharp$,
Dong Li$^\sharp$,
Weigao Sun$^\sharp$,
Weixuan Sun$^\sharp$,
Xuyang Shen$^\sharp$,
}\\
{\normalsize
\textbf{
 Xiaodong Han, Yunshen Wei, Baohong Lv, Xiao Luo, Yu Qiao,
 Yiran Zhong\thanks{Yiran Zhong is the corresponding author. Email: \texttt{zhongyiran@gmail.com}. $\sharp$ equal contribution.}
 }
}\\
\hspace{0.1mm} 
 OpenNLPLab, Shanghai AI Laboratory\\
\hspace{0.1mm}  \texttt{https://github.com/OpenNLPLab/TransnormerLLM} 
}

\iclrfinalcopy 
\begin{document}

\maketitle

\begin{abstract}
We present TransNormerLLM, the first linear attention-based Large Language Model (LLM) that outperforms conventional softmax attention-based models in terms of both accuracy and efficiency. TransNormerLLM evolves from the previous linear attention architecture TransNormer~\citep{qin-etal-2022-devil} by making advanced modifications that include positional embedding, linear attention acceleration, gating mechanism, tensor normalization, and inference acceleration and stabilization. 
Specifically, we use LRPE~\citep{qin2023linearized} together with an exponential decay to avoid attention dilution issues while allowing the model to retain global interactions between tokens.
Additionally, we propose Lightning Attention, a cutting-edge technique that accelerates linear attention by more than twice in runtime and reduces memory usage by a remarkable four times.
To further enhance the performance of TransNormer, we leverage a gating mechanism to smooth training and a new tensor normalization scheme to accelerate the model, resulting in an impressive acceleration of over $20\%$. 
Furthermore, we develop a robust inference algorithm that ensures numerical stability and consistent inference speed, regardless of the sequence length, showcasing superior efficiency during both training and inference stages.
We also implement an efficient model parallel schema for TransNormerLLM, enabling seamless deployment on large-scale clusters and facilitating expansion to even more extensive models, \ie LLMs with 175B parameters. We validate our model design through a series of ablations and train models with sizes of 385M, 1B, and 7B on our self-collected corpus. Benchmark results demonstrate that our models not only match the performance of state-of-the-art LLMs with Transformer but are also significantly faster.
\end{abstract}

\section{Introduction}
The field of Natural Language Processing (NLP) has been revolutionized by the advent of large-scale language models (LLMs)~\citep{touvron2023llama, biderman2023pythia,brown2020language}. These models have demonstrated exceptional performance across a multitude of tasks, elevating abilities to comprehend, generate, and interact with human languages in computational frameworks. Previous language modeling development has predominantly centered around Transformer architectures, with seminal models such as vanilla Transformer~\citep{vaswani2017attention}, GPT series~\citep{radford2018improving,radford2019language, brown2020language}, BERT~\citep{devlin-etal-2019-bert}, and BART~\citep{lewis2019bart} standing as standard backbones in related fields. The success of Transformer architectures is premised on the softmax attention mechanism, which discerns dependencies among input tokens in a data-driven scheme and has global position awareness, offering the model an effective way to handle the long-range dynamism of natural language.

Nevertheless, conventional Transformers are not without their constraints. Primarily, their quadratic time complexity with respect to the sequence length limits their scalability and hampers efficiency in terms of computational resources and time during the training and inference stages. Numerous efficient sequence modeling methods have been proposed in an attempt to reduce the quadratic time complexity to linear~\citep{katharopoulos2020transformers,choromanski2021rethinking,zhen2022cosformer,zheng2023efficient,zheng2022linear}. However, there are two reasons that prohibit them to be applied to LLMs: 1) their performance in language modeling is often unsatisfactory; 2) they do not demonstrate speed advantages in real-world scenarios.

In this paper, we introduce TransNormerLLM, the first linear attention-based LLM that surpasses conventional softmax attention in both accuracy and efficiency. The development of TransNormerLLM builds upon the foundations of the previous linear attention architecture, TransNormer~\citep{qin-etal-2022-devil}, while incorporating a series of advanced modifications to achieve superior performance. The key enhancements in TransNormerLLM include positional embedding, linear attention acceleration, gating mechanism, tensor normalization, and inference acceleration. 

One notable improvement is the replacement of the TransNormer's DiagAttention with Linear Attention to enhance global interactions. To address the issue of dilution, we introduced LRPE~\citep{qin2023linearized} with exponential decay~\citep{alibi,qin2023toeplitz,2305.13048}. Lightning Attention, a novel technique that significantly accelerates linear attention during training is introduced, resulting in a more than two-fold improvement, while also reducing memory usage by four times with IO awareness. Furthermore, we simplified GLU and Normalization, with the latter leading to a 20\% speedup. A robust inference algorithm ensures the stability of numerical values and constant inference speed, regardless of the sequence length, thereby enhancing the efficiency of our model during both training and inference stages.

We validate the efficacy of TransNormerLLM on our self-collected pre-train corpus, which is more than $6$TB in size and contains over $2$ trillion tokens. We expand the original TransNormer model, ranging from 385M to 175B parameters, and benchmark models with sizes of 385M, 1B, and 7B. The benchmark results demonstrate that our models achieve competitive performance with existing state-of-the-art transformer-based LLMs with similar sizes while also having faster inference speeds. We will open-source our pre-trained models, enabling researchers and practitioners to build upon our work and explore efficient transformer structures in LLMs.

\section{Related Work}
\subsection{Transformer-based LLMs}
In recent years, the field of Large Language Models (LLMs) has experienced significant advancements. Adhering to the scaling laws~\citep{kaplan2020scaling}, various LLMs with over 100 billion parameters have been introduced, such as GPT-3~\citep{brown2020language}, Gopher~\citep{rae2022scaling}, PaLM~\citep{2204.02311}, GLM~\citep{du2022glm} and \emph{etc.}. More specialized models like Galactica~\citep{taylor2022galactica} have also emerged for specific domains like science.
A notable development is Chinchilla~\citep{hoffmann2022training}, an LLM model with 70 billion parameters that redefines these scaling laws, focusing on the number of tokens rather than model weights. Furthermore, LLaMA~\citep{touvron2023llama} has also sparked interest due to its promising performance and open-source availability.
The discourse around LLMs also encompasses the dynamics between open-source and closed-source models. Open-source models such as BLOOM~\citep{workshop2023bloom}, OPT~\citep{zhang2022opt}, LLaMA~\citep{touvron2023llama}, Pythia~\citep{biderman2023pythia} and Falcon~\citep{penedo2023refinedweb} are rising to compete against their closed-source counterparts, including GPT-3~\citep{brown2020language} and Chinchilla~\citep{hoffmann2022training}. To speed up training, Sparse Attention~\citep{1904.10509,beltagy2020longformer} was introduced, but among large models, only GPT-3 adopted it~\citep{brown2020language,2210.15424}.

\vspace{-3mm}
\subsection{Non-Transformer-based LLMs Candidates}
Despite the proliferation of Transformer-based large models in the research community, a portion of recent work has prioritized addressing its square time complexity. This focus has led to the exploration and development of a series of model architectures that diverge from the traditional Transformer structure. Among them, four significant contenders—linear transformers, state space model, long convolution, and linear recurrence—have shown promising results as substitutes for self-attention (SA) modules when modeling long sequences. These alternatives are favored for their superior asymptotic time complexity and competitive performances.

\vspace{-3mm}
\paragraph{Linear Transformer}
Linear Transformer decomposes Softmax Attention into the form of the inner product of hidden representations, which allows it to use the "Right Product Trick," where the product of keys and values is computed to avoid the quadratic $n \times n$ matrix. Different methods utilize various hidden representations. For example, ~\citet{katharopoulos2020transformers} use 1+elu as an activation function, ~\citet{zhen2022cosformer} use the cosine function to approximate the properties of softmax, and ~\citet{ke2021rethinking,zheng2022linear,zheng2023efficient} approximate softmax through theoretical approaches. Although its theoretical complexity is $O(nd^2)$, the actual computational efficiency of Linear Attention becomes quite low when used in causal attention due to the need for \textit{cumsum} operations~\citep{hua2022transformer}. On the other hand, most Linear Transformers still exhibit a certain performance gap compared to traditional Transformers~\citep{katharopoulos2020transformers,liu2022neural}.

\vspace{-3mm}
\paragraph{State Space Model}
State Space Model is based on the State Space Equation for sequence modeling~\citep{s4}, using special initialization~\citep{2008.07669,s4d}, diagonalization assumptions~\citep{gupta2022DSS}, and some techniques~\citep{h3} to achieve performance comparable to Transformers. On the other hand, due to the characteristics of the State Space Equation, it enables inference to be conducted within constant complexity~\citep{s4}.

\vspace{-3mm}
\paragraph{Long Convolution}
Long convolution models~\citep{qin2023toeplitz,simplelongconv} utilize a kernel size equal to the input sequence length, facilitating a wider context compared to traditional convolutions. Training these models involves the efficient $O(n\log n)$ Fast Fourier Transforms (FFT) algorithm. However, long convolutions pose certain challenges, such as the need for causal convolution inference, which necessitates caching all historical computations similar to SA's key-value (KV) cache. The memory requirements for handling long sequences, coupled with the higher inference complexity compared to RNNs, make them less ideal for processing long sequences.
\paragraph{Linear RNN}
\vspace{-3mm}
Linear RNNs~\citep{2303.06349,peng2023rwkv}, in contrast, stand out as more suitable replacements for SA in long-sequence modeling. A notable example is the RWKV ~\citep{peng2023rwkv} model, a linear RNN-based LLM that has shown competitive performance against similarly scaled GPT models.

\section{TransNormerLLM}
\subsection{Architecture Improvement}
In this section, we thoroughly investigate each module of the network and propose several improvements to achieve an optimal balance between efficiency and performance. Below, we outline the key designs of each block along with the inspiration behind each change. For the details of configurations for TransNormerLLM variants from 385M to 175B parameters, see Appendix~\ref{app:model}.

\subsubsection{Improvement 1: Position encoding}
In TransNormer, DiagAttention is used at the lower layers to avoid dilution issues. However, this leads to a lack of global interaction between tokens. 
In TransNormerLLM, we leverage LRPE~\citep{qin2023linearized} with exponential decay~\citep{alibi,qin2023toeplitz,peng2023rwkv} to address this issue, retaining full attention at the lower layers. The expression of our position encoding is as follows:
\begin{equation}
\small
a_{st}=\mathbf q_s^{\top} \mathbf k_t \lambda^{s-t}\exp^{i\theta(s-t)}.
\label{eq: pe}
\end{equation}
which we call LRPE-d - Linearized Relative Positional Encoding with exponential decay. Similar to the original LRPE, we set $\theta$ to be learnable. We empirically find that rather than applying LRPE-d to every layer, applying it to the first layer and keeping other layers with exponential decay can speed up training by approximately 15-20\% but only with a subtle effect on the performance.

Note that this position encoding is fully compatible with Linear Attention, as it can be decomposed with respect to $s$ and $t$ separately.
The value of $\lambda$ for the $h$-th head in the $l$-th layer (assuming there are a total of $H$ heads and $L$ layers) is given by:
\begin{equation}
\label{eq:decay}
\small
\textstyle
\lambda =\exp\left(-\frac{8h}{H}\times \left(1-\frac{l}{L}\right) \right).
\end{equation}
Here, $\frac{8h}{H}$ corresponds to the decay rate of the $h$-th head, while $ \left(1-\frac{l}{L}\right)$ corresponds to the decay rate of the $l$-th layer. The term $ \left(1-\frac{l}{L}\right)$ ensures that the Theoretical Receptive Fields (TRF)~\citep{2307.10156} at the lower layers is smaller compared to the higher layers, which aligns with TransNormer's motivation. It should be noted that the decay rate in the last layer is set to 1, allowing each token to attend to global information. We choose $\lambda$ to be non-learnable since we empirically found that gradients become unstable when $\lambda$ is learnable, leading to NaN values.

\subsubsection{Improvement 2: Gating mechanism}
Gate can enhance the performance of the model and smooth the training process. In TransNormerLLM, we adopted the approach from Flash~\citep{hua2022transformer} and used the structure of Gated Linear Attention (GLA) in token mixing:
\begin{equation}
\small
\mathrm{TokenMixer}: \mathbf{O}=\mathrm{Norm}(\mathbf{Q} \mathbf{K}^{\top}\mathbf{V})\odot \mathbf{U},
\label{eq: gla1}
\end{equation}
\vspace{-4mm}
where:
\begin{equation}
\small
\mathbf Q=\phi(\mathbf X \mathbf W_q),\mathbf K=\phi(\mathbf X \mathbf W_k),\mathbf V=\mathbf X \mathbf W_v,\mathbf U=\mathbf X \mathbf W_u.
\label{eq: gla2}
\end{equation}

We choose $\phi$ to be swish~\citep{1710.05941} activation function as we empirically find that it outperforms other activation functions, as shown in Table~\ref{tab:gla_act}.


To further accelerate the model, we propose Simple GLU (SGLU), which removes the activation function from the original GLU structure as the gate itself can introduce non-linearity. Therefore, our channel mixing becomes:
\begin{equation}
\vspace{-1mm}
\small
\mathrm{ChannelMixer}:\mathbf {O}=[\mathbf V\odot \mathbf U]\mathbf W_o,\\
\mathbf V=\mathbf X \mathbf W_v,\mathbf U=\mathbf X \mathbf W_u,
\label{eq: glu}
\end{equation}
We empirically find that not using an activation function in GLU will not lead to any performance loss, as demonstrated in Table~\ref{tab:glu_act}.

\subsubsection{Improvement 3: Tensor normalization}
We employ the NormAttention introduced in TransNormer~\citep{qin-etal-2022-devil} as follows:
\begin{equation}
\small
\mathbf{O}=\mathrm{Norm}((\mathbf{Q} \mathbf{K}^{\top})\mathbf{V})
\label{eq: norm attention}
\end{equation}
This attention mechanism eliminates the softmax and scaling operation. Moreover, it can be transformed into linear attention through right multiplication:
\begin{equation}
\small
\mathbf{O}=\mathrm{Norm}(\mathbf{Q} (\mathbf{K}^{\top}\mathbf{V}))
\label{eq: norm attention 2}
\end{equation}
This linear form allows for recurrent prediction with a complexity of $O(nd^2)$, making it efficient during inference. Specifically, we only update $\mathbf{K}^{\top}\mathbf{V}$ in a recurrent manner without computing the full attention matrix.
In TransNormerLLM, we replace the RMSNorm with a new simple normalization function called SimpleRMSNorm, abbreviated as SRMSNorm:
\begin{equation}
\small
\textstyle
\mathrm{SRMSNorm}(\mathbf x)=\frac{\mathbf x}{\|\mathbf x \|_2/\sqrt d}.
\end{equation}
We empirically find that using SRMSNorm does not lead to any performance loss, as demonstrated in the ablation study in Table.~\ref{tab:norm}.

\begin{wrapfigure}[18 ]{r}{0.5\textwidth}
  \centering
  \vspace{-17mm}
    \includegraphics[width=0.5\textwidth]{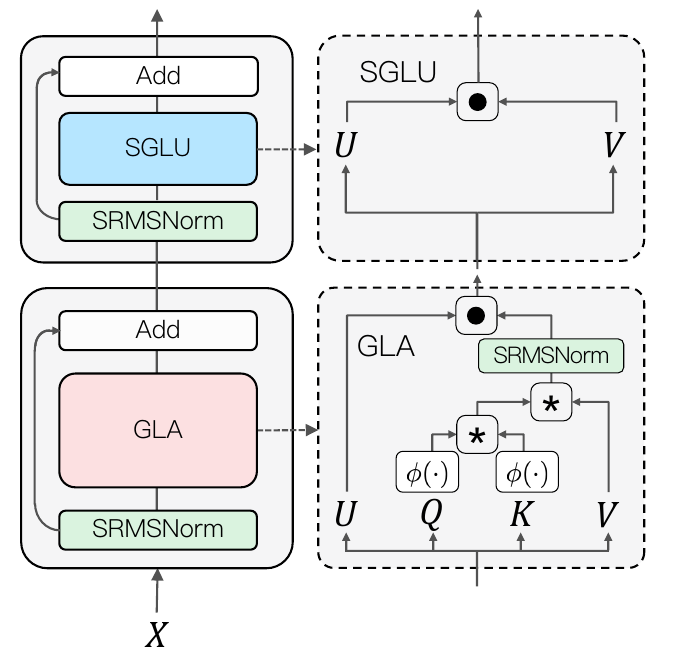}
    \vspace{-8mm}
  \captionof{figure}{Architecture overview of the proposed model. 
  Each transformer block is composed of a Gated Linear Attention(GLA) for token mixing and a Simple Gated Linear Unit (SGLU) for channel mixing.  
  We apply pre-norm for both modules.}  
  \label{fig:arch}
  
\end{wrapfigure}

\subsubsection{The overall structure}
The overall structure is illustrated in Figure~\ref{fig:arch}.
In this structure, the input $\mathbf X$ is updated through two consecutive steps: First, it undergoes Gated Linear Attention (GLA) with the application of SimpleRMSNorm (SRMSNorm) normalization. Then, it goes through the Simple Gated Linear Unit (SGLU) with SRMSNorm normalization again. This overall architecture helps improve the model's performance based on the PreNorm approach. The pseudo-code of the overall process is as follows:
\small
\begin{equation}
\begin{gathered}
\mathbf X = \mathbf X + \mathrm{GLA}(\mathrm{SRMSNorm}(\mathbf X)), \\
\mathbf X = \mathbf X + \mathrm{SGLU}(\mathrm{SRMSNorm}(\mathbf X)).
\end{gathered}
\end{equation}
\normalsize

\subsection{Training Optimization}
\subsubsection{Lightning Attention}
The structure of linear attention allows for efficient attention calculation with a complexity of $O(nd^2)$ through right-multiplication. However, for causal prediction, right-multiplication is not efficient as it necessitates \textit{cumsum} computation~\citep{hua2022transformer}, which hinders parallelism training. 
As a result, during training, we continue to use the conventional left-multiplication version.
To accelerate attention calculations, we introduce the Lightning Attention algorithm inspired by~\citep{dao2023flashattention2,dao2022flashattention}, which makes our linear attention IO-friendly.
It computes the following:
\begin{equation}
\small
\mathbf O= (\mathbf Q\mathbf K^{\top} \odot \mathbf M)\mathbf V.
\end{equation}
Here, $\mathbf M$ is the attention mask which enables lower triangular causal masking and positional encoding.
In the Lightning Attention, we split the inputs $\mathbf{Q}, \mathbf{K}, \mathbf{V}$ into blocks, load them from slow HBM to fast SRAM, then compute the attention output with respect to those blocks. Then we accumulate the final results.
The computation speed is accelerated by avoiding the operations on slow HBM.
The implementation details of Lightning Attention are shown in Appendix~\ref{app:lightning}, where Algorithm~\ref{algo:lightning attention fw pseudo} for forward pass and Algorithm~\ref{algo:lightning attention bw pseudo} for backward pass. 

\subsubsection{Model Parallelism on TransNormerLLM}

To effectively execute large-scale pre-training for TransNormerLLM, we have put efforts on system optimization encompassing various dimensions. Specifically, we employ fully sharded data parallelism (FSDP)~\citep{zhao2023pytorch}, a technique that shards all model parameters, gradients, and optimizer state tensors across the entire cluster. This strategic partition significantly reduces the memory footprint on each individual GPU, thereby enhancing memory utilization.
In our pursuit of greater efficiency, we leverage activation checkpointing~\citep{shoeybi2019megatron}, which minimizes the cached activations in memory during the forward pass. Instead of retaining these activations, they are recomputed when calculating gradients in the backward pass. This approach saves huge GPU memory thus enable to apply bigger batch size.
Furthermore, we harness automatic mixed precision (AMP)~\citep{micikevicius2017mixed} to simultaneously save GPU memory and expedite computational speed. It's noteworthy that in our experimental setup, we employ BFloat16~\citep{kalamkar2019study} due to its observed advantage in enhancing the training stability of TransNormerLLM models.

In addition to the previously mentioned optimization endeavors, we delve deeper into the realm of system engineering by implementing model parallelism specifically tailored to linear transformers, drawing inspiration from Megatron-LM model parallelism~\citep{shoeybi2019megatron}.
In a standard transformer model, each transformer layer comprises a self-attention block followed by a two-layer multi-layer perceptron (MLP) block. Megatron-LM model parallelism independently addresses these two constituent blocks. Similarly, within the architecture of TransNormerLLM, characterized by its two primary components, SGLU and GLA, we apply model parallelism to each of these components separately. The intricate details of our model parallelism strategies are elaborated below.

\paragraph{Model Parallelism on SGLU}
Recall the SGLU structure in (\ref{eq: glu}):
\begin{equation}
\small
\mathbf O=[(\mathbf X \mathbf W_v) \odot (\mathbf X \mathbf W_u)]\mathbf W_o,
\label{eq: mp_glu}
\end{equation}
The model parallelism adaptation of SGLU is as follows:
\begin{equation}
\small
[\mathbf {O}'_1, \mathbf {O}'_2]=\mathbf X[\mathbf W_v^1, \mathbf W_v^2] \odot \mathbf X[\mathbf W_u^1, \mathbf W_u^2],
=[\mathbf X \mathbf W_v^1, \mathbf X \mathbf W_v^2] \odot [\mathbf X \mathbf W_u^1, \mathbf X \mathbf W_u^2],
\label{eq: mp_glu_1}
\end{equation}
which splits the weight matrices $\mathbf W_v$ and $\mathbf W_u$ along their columns and obtains an output matrix splitting along its columns too.
Then the split output $[\mathbf O_1, \mathbf O_2]$ is multiplied by another matrix which is split along its rows as:
\begin{equation}
\small
\mathbf {O}=[\mathbf O_1', \mathbf O_2'] [\mathbf W_o^1, \mathbf W_o^2]^\top=\mathbf O_1' \mathbf W_o^1 + \mathbf O_2' \mathbf W_o^2
\label{eq: mp_glu_output}
\end{equation}
Similar with model parallelism in Megatron-LM, this whole procedure splits three general matrix multiplies (GEMMs) inside the SGLU block across multiple GPUs and only introduces a single \textit{all-reduce} collective communication operation in both the forward and backward passes, respectively. 

\paragraph{Model Parallelism on GLA}
Recall the GLA block in (\ref{eq: gla1}) and (\ref{eq: gla2}), its model parallelism version is:
\begin{equation}
\small
[\mathbf{O_1}, \mathbf{O_2}]=\mathrm{SRMSNorm}(\mathbf{Q} \mathbf{K}^{\top}\mathbf{V})\odot \mathbf{U},
\label{eq: mp_gla1}
\end{equation}
where:
\small
\begin{align}
\mathbf Q=[\phi(\mathbf X \mathbf W_q^1), \phi(\mathbf X \mathbf W_q^2)],
\mathbf K=[\phi(\mathbf X \mathbf W_q^1), \phi(\mathbf X \mathbf W_q^2)],
\mathbf V=\mathbf X [\mathbf W_v^1, \mathbf W_v^2],\mathbf U=\mathbf X [\mathbf W_u^1, \mathbf W_u^2],
\label{eq: mp_gla2}
\end{align}
\normalsize
Note that in our implementation, we use the combined QKVU projection to improve computation efficiency for linear attention.
The obtained split output matrix $[\mathbf{O_1}, \mathbf{O_2}]$ again is multiplied by a weight matrix split along its columns which is similar to (\ref{eq: mp_glu_output}).

\subsection{Robust Inference}
In this section, we discuss the inference problem in TransNormerLLM. It is important to note that the formula~\ref{eq: pe} can be decomposed into the following form:
\vspace{-1mm}
\begin{equation}
\small
a_{st}=(\mathbf q_s \lambda^s \exp^{i\theta s})^{\top}(\mathbf k_t \lambda^{-t} \exp^{i\theta t}).
\end{equation}
This allows TransNormerLLM to perform inference in the form of an RNN. Details of the procedure are shown in Algorithm~\ref{algo:origin}. However, it is worth noting that $\lambda < 1$, which results in:
\vspace{-1mm}
\begin{equation}
\small
\|\mathbf q_s \lambda^s \exp^{i\theta s}\|_2= \|\mathbf q_s\|_2 \lambda^s \to 0,\\
\|\mathbf k_t \lambda^{-t} \exp^{i\theta t}\|_2= \|\mathbf k_t\|_2 \lambda^{-t} \to \infty,
\end{equation}
leading to numerical precision issues.

\vspace{-2mm}
To avoid these issues, we propose a Robust Inference Algorithm in~\ref{algo:robust}. Since $\|\mathbf q_s \exp^{i\theta s} \|=\|\mathbf q_s \|$, $\|\mathbf k_t \exp^{i\theta t} \|=\|\mathbf k_t \|$, for simplicity, we will omit LRPE~\citep{qin2023linearized} in the subsequent discussions, considering only $a_{st}=\mathbf q_s^{\top} \mathbf k_t \lambda^{s-t} .$ We provide a mathematical proof of $[\mathbf {kv}]_t=\lambda^{-t}[{\mathbf {\overline{kv}}}]_t$
in Appendix~\ref{app:robustinfer}
\small
\begin{figure}[t]
\vspace{-18mm}
    \begin{minipage}{.48\textwidth}
        \centering       
        \begin{algorithm}[H]
            \caption{Origin Inference Algorithm}
            \label{algo:origin}
            \begin{algorithmic}
            \State{\textbf{Input:} $\mathbf q_t, \mathbf k_t, \mathbf v_t, t=1,\ldots, n$; }
            \State{\textbf{Output:} $\mathbf o_t, t=1,\ldots, n$;}
             \State{\textbf{Initialize:} 
                $[\mathbf {kv}]_0=\mathbf 0$;
             }
            \For{${t=1,\ldots, n}$}
                \State {
                $[\mathbf {kv}]_t = [\mathbf {kv}]_{t-1} + \mathbf {k_t} \lambda^{-t} \mathbf {v}_t^{\top}$,
                }
                \State {
                $\mathbf o_t=\mathbf {q}_t \lambda^t [\mathbf {kv}]_t$.
                }
              \EndFor
            \end{algorithmic}
        \end{algorithm}
    \end{minipage}%
    \hspace{4mm}
    \begin{minipage}{0.48\textwidth}
    \begin{algorithm}[H]
        \caption{Robust Inference Algorithm}
        \label{algo:robust}
        \begin{algorithmic}
        \State{\textbf{Input:} $\mathbf q_t, \mathbf k_t, \mathbf v_t, t=1,\ldots, n$; }
        \State{\textbf{Output:} $\mathbf o_t, t=1,\ldots, n$;}
         \State{\textbf{Initialize:} 
            $[\mathbf {\overline{kv}}]_0=\mathbf 0$;
         }
        \For{${t=1,\ldots, n}$}
            \State {
            $[\mathbf {\overline{kv}}]_t = \lambda [\mathbf {\overline{kv}}]_{t-1} + \mathbf {k_t}  \mathbf {v}_t^{\top}$,
            }
            \State {
            $\mathbf o_t=\mathbf {q}_t [\mathbf {\overline{kv}}]_t$.
            }
          \EndFor
    \end{algorithmic}
    \end{algorithm}
    \end{minipage}%
\end{figure}
\normalsize

\section{Experiments}
\label{section: experiment}
We use PyTorch~\citep{NEURIPS2019_9015} and Triton~\citep{Tillet2019TritonAI} to implement TransNormerLLM in Metaseq framework~\citep{zhang2022opt}. Our model is trained using Adam optimizer~\citep{kingma2017adam}, and we employ FSDP to efficiently scale our model to NVIDIA A100 80G clusters. We additionally leverage the model parallel as appropriate to optimize performance. In ablation studies, all models are trained on a sampled corpus from our corpus with 300B tokens. In order to reduce the fluctuation of Losses and PPLs in the tables below, we compute the average Losses and PPLs of the last 1k iterations as the final metrics. For our benchmark models, we train our 385M, 1B, and 7B models on our corpus for 1 trillion, 1.2 trillion, and 1.4 trillion tokens respectively. We use an input sequence length of 8192 tokens in our pretraining process.
For a comprehensive understanding of our corpus, encompassing intricate details such as data preprocessing methods and tokenization procedures, we direct interested readers to Appendix~\ref{app:corpus}.

\subsection{Architecture Ablations}
\paragraph{Transformer \emph{vs} TransNormerLLM} 
We carried out a meticulous series of comparative tests between our TransNormerLLM and Transformer, spanning over an array of disparate sizes. The comparative performance of these models is clearly illustrated in Table~\ref{tab:transf_vs_transn}. 
Under identical configurations, it becomes evident that our TransNormerLLM exhibits a superior performance profile compared to Transformer. We observed that TransNormerLLM outperformed Transformer by a remarkable 5\% at the size of 385M. More importantly, as the size reached 1B, this superiority became even more pronounced, with an advantage of 9\% for TransNormerLLM over Transformer.

\begin{table}[h]
\centering
    \small
    \caption{\small\textbf{Transformer \emph{vs} TransNormerLLM.} TransNormerLLM performs better than Transformer in size of 385M and 1B under identical configurations by 5\% and 9\%, respectively.}
    \label{tab:transf_vs_transn}
    \vspace{-3mm}
    \setlength{\tabcolsep}{4.6mm}
    \begin{tabular}{cccccccc}
    \hline
    \small
    \\[-1em]
    \multicolumn{2}{c}{Model Size}  & \multicolumn{3}{c}{385M}  & \multicolumn{3}{c}{1B}               \\ 
    \cmidrule(lr){1-2} \cmidrule(lr){3-5} \cmidrule(lr){6-8}  
    \\[-1em]
    \multicolumn{2}{c}{Method}      & \multicolumn{1}{c}{Updates} & Loss & PPL  & \multicolumn{1}{c}{Updates} & Loss& PPL   \\ \hline
    \\[-1em]
    \multicolumn{2}{c}{Transformer} & 100K                         & 2.362 &5.160 & 100K                         & 2.061 &4.765 \\ 
    \multicolumn{2}{c}{TransNormerLLM} & 100K                         & 2.248 &4.770 & 100K                         & 1.896 &3.729\\ \hline
\end{tabular}
\end{table}

\begin{wraptable}[5]{r}{.5\linewidth}
    \centering
  \small
  \vspace{-5mm}
    \caption{\small\textbf{TransNormer \emph{vs} TransNormerLLM.}}
    \label{tab:transnormers}
    \centering
    \setlength{\tabcolsep}{1.2mm}
    \vspace{-4mm}
        \begin{tabular}{lllll}
        \hline
        \\[-1em]
        Method & Params & Updates & Loss & PPL \\ \hline
        \\[-1em]
        TransNormerLLM & 385M & 100K  & 2.248 &4.770 \\ 
        TransNormer-T1 & 379M & 100K & 2.290 &4.910 \\ 
        TransNormer-T2 & 379M & 100K & 2.274 &4.858\\ \hline
    \end{tabular}
\end{wraptable}

\paragraph{TransNormer \emph{vs} TransNormerLLM} 
We compare the original TransNormer and the improved TransNormerLLM and the results are shown in Table~\ref{tab:transnormers}. TransNormerLLM exhibited an enhancement of 2\% and 1\% respectively.

\begin{wraptable}[9]{r}{.5\linewidth}
    \centering
  \small
    \caption{\textbf{Positional encoding.} LRPE-d leads to the most optimal outcome.}
    \vspace{-2mm}
    \label{tab:pe}
    \centering
    \setlength{\tabcolsep}{2mm}
     \begin{tabular}{lllll}
    \hline
        PE Methods & Params & Updates & Loss &PPL\\ \hline
         Mix & 385M  & 100K& 2.248 &4.770\\ 
            APE & 386M & 100K & 2.387 & 5.253\\
            Exp-Decay & 385M & 100K & 2.267 & 4.834 \\ 
            LRPE & 385M & 100K & 2.287 & 4.899 \\             
            LRPE-d & 385M  & 100K & 2.236 &4.728  \\ 
       \hline
    \end{tabular}
\end{wraptable}

\paragraph{Positional Encoding} In the positional encoding experiment, we conducted a series of tests, comparing Mix (LRPE-d for the first layer, Exp-Decay for the rest), APE (Absolute Positional Encoding), LRPE, Exp-Decay (Exponential Decay), and LRPE-d. As evident from Table~\ref{tab:pe}, 
Ours and LRPE-d achieve better performance than other options. We select the Mix positional encoding as it boosts the training speed up to 20\% while only slightly worse than LRPE-d.

\begin{wraptable}[4]{r}{.5\linewidth}
    \centering
  \small
  \vspace{-8.2mm}
    \caption{\textbf{Ablations on decay temperature.} The results of decay temperature proved to be superior.}
    \label{tab:temperature}
    \centering
    \vspace{-2mm}
    \setlength{\tabcolsep}{1.6mm}
        \begin{tabular}{lllll}
        \hline
            Temperature & Params & Updates & Loss &PPL \\ \hline
            w/ temperature &  385M & 100K& 2.248 &4.770\\ 
            w/o temperature & 385M & 100K & 2.258 &4.804 \\ \hline
    \end{tabular}
\end{wraptable}
We also perform ablations on the decay temperature $\left(1-\frac{l}{L}\right)$ in Eq.~\ref{eq:decay}. The perplexity of the TransNormerLLM is reduced by adding the decay temperature, as shown in Table~\ref{tab:temperature}.

\vspace{-2.5mm}
\begin{wraptable}[5]{r}{.5\linewidth}
    \centering
  \small
  \vspace{-4.5mm}
    \caption{\textbf{Ablations on gating mechanism.} The performance with the gate proved to be superior.}
    \vspace{-2mm}
    \label{tab:gate}
    \centering
    \setlength{\tabcolsep}{2.5mm}
        \begin{tabular}{lllll}
        \hline
            Gate & Params & Updates & Loss &PPL\\ \hline
            w/ gate & 385M & 100K & 2.248 &4.770\\ 
            w/o gate & 379M & 100K & 2.263 &4.820 \\ \hline
    \end{tabular}
\end{wraptable}

\paragraph{Gating Mechanism} We conduct ablation studies to examine the effect of including the gating mechanism. As observed in Table~\ref{tab:gate}, gate enabled the reduction of the loss value from 2.263 to 2.248.

\vspace{-5mm}
\begin{wraptable}[7]{r}{.5\linewidth}
    \centering
  \small
  \vspace{-2mm}
    \caption{\textbf{Ablations on GLA activation functions.} The results obtained from different activation functions were virtually identical.}
    \vspace{-2mm}
    \label{tab:gla_act}
    \centering
    \setlength{\tabcolsep}{2.4mm}
         \begin{tabular}{lllll}
        \hline
        GLA Act & Params & Updates & Loss &PPL \\ \hline
        Swish & 385M & 100K & 2.248 &4.770\\
        No Act & 385M & 100K & 2.283 &4.882 \\ 
        1+elu	& 385M	& 100K	& 2.252	& 4.767 \\
        \hline
    \end{tabular}
\end{wraptable}
\paragraph{GLA Activation Functions} 
We conducted experiments on the GLA (Gated Linear Attention) structure with respect to the activation function. As shown in Table ~\ref{tab:gla_act}, using Swish and 1+elu leads to similar performance. However, in our experiments, using 1+elu in our 7B model may encounter a NaN problem, so we use Swish in our model.

\begin{wraptable}[7]{r}{.5\linewidth}
    \centering
  \small
  \vspace{-4.5mm}
    \caption{\textbf{Ablations on GLU activation functions.} The exclusion of the activation function had no negative impact on the results.}
    \vspace{-2mm}
    \label{tab:glu_act}
    \centering
    \setlength{\tabcolsep}{2.4mm}
         \begin{tabular}{lllll}
        \hline
        GLU Act & Params & Updates & Loss &PPL \\ \hline
        No Act & 385M   & 100K & 2.248 &4.770 \\  
        Swish & 385M & 100K & 2.254 & 4.788\\ \hline
    \end{tabular}
\end{wraptable}

\paragraph{GLU Activation Functions} 
We conduct an experiment by removing the activation function within the Gated Linear Units (GLU) structure. As shown in Table~\ref{tab:glu_act}, the results reveal that this alteration had a negligible impact on the final outcome. As a result, we decide to adopt the Simple Gated Linear Units (SGLU) structure in our final model configuration.

\vspace{-4mm}
\begin{wraptable}[8]{r}{.5\linewidth}
    \centering
  \small
  \vspace{-4mm}
    \caption{\textbf{Normalization Functions.} The deviation in results among the bellowing normalization functions is minimal.}
    \label{tab:norm}
    \centering
    \vspace{-2mm}
    \setlength{\tabcolsep}{2mm}
     \begin{tabular}{lllll}
    \hline
        Norm Type & Params & Updates & Loss &PPL \\ \hline
        SRMSNorm & 385M  & 100K & 2.248 &4.770 \\  
        RMSNorm & 385M & 100K & 2.247  & 4.766\\ 
        LayerNorm & 385M & 100K & 2.247 &4.765 \\ \hline
    \end{tabular}
\end{wraptable}

\begin{figure}[htb]
    \centering
    \includegraphics[width=0.9\textwidth]{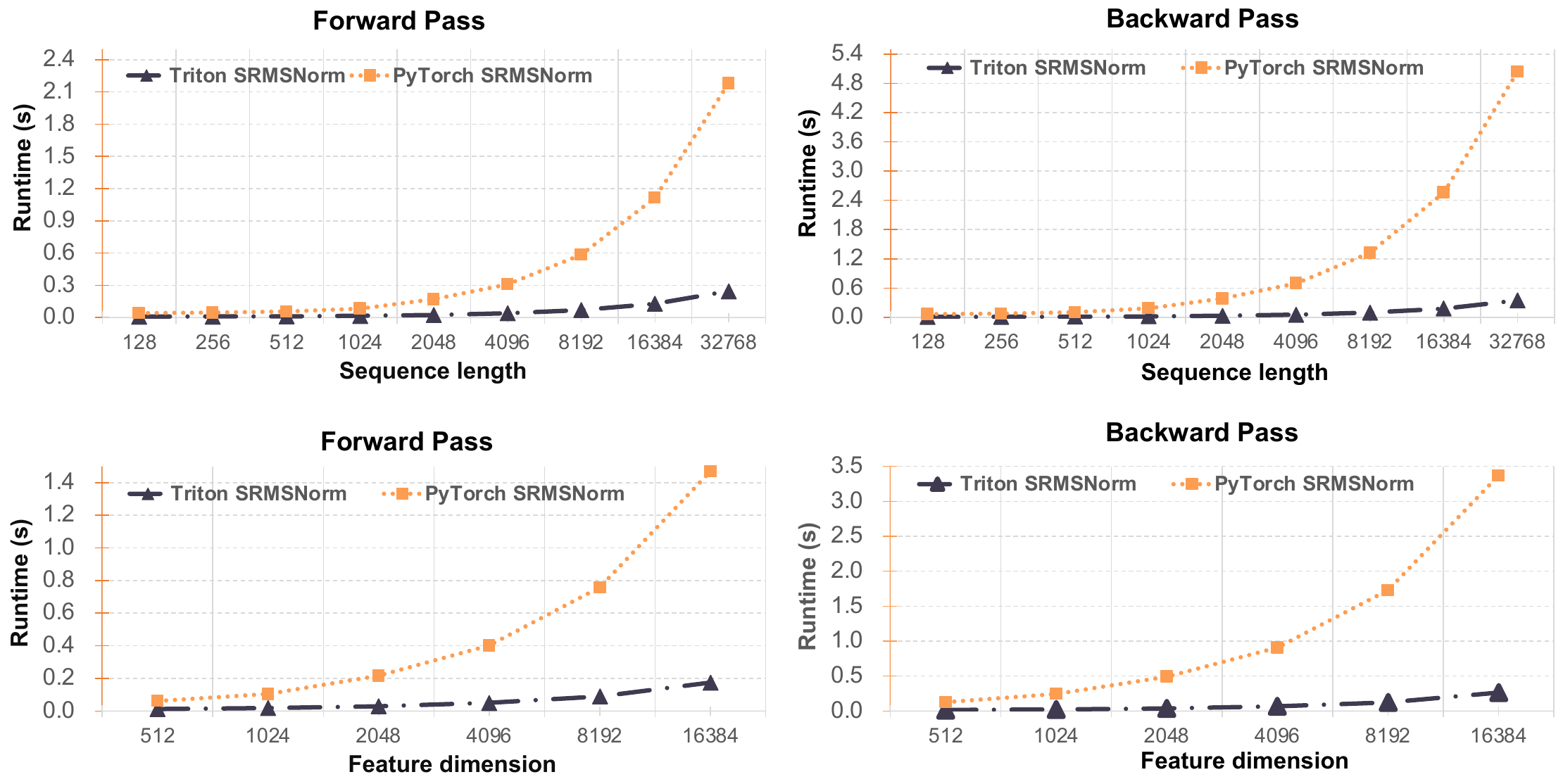}
    \vspace{-2mm}
    \caption{\textbf{Performance Evaluation of SRMSNorm Implementation.} The upper figures exhibit the runtime comparison of the forward pass (left) and backward pass (right) for different sequence lengths, with a fixed feature dimension of 3072. The lower two figures illustrate the runtime comparison for various feature dimensions, with a fixed sequence length of 4096.}
    \vspace{-4mm}
    \label{fig:norm}
\end{figure}

\paragraph{Normalization functions} In our study, we conducted a series of ablation tests employing various normalization methods including SRMSNorm, RMSNorm and LayerNorm. The results indicate that there is almost no difference among these methods when applied to TransNormerLLM. Nevertheless, during the course of our testing, we revisited and re-engineered the SRMSNorm using Triton. As it is shown in Figure ~\ref{fig:norm}, empirical evidence supports that our modification offers a significant boost in computational speed when operating with larger dimensions, compared to the PyTorch implementation methods.

\begin{figure}[htb]
\centering
\includegraphics[width=0.9\textwidth]{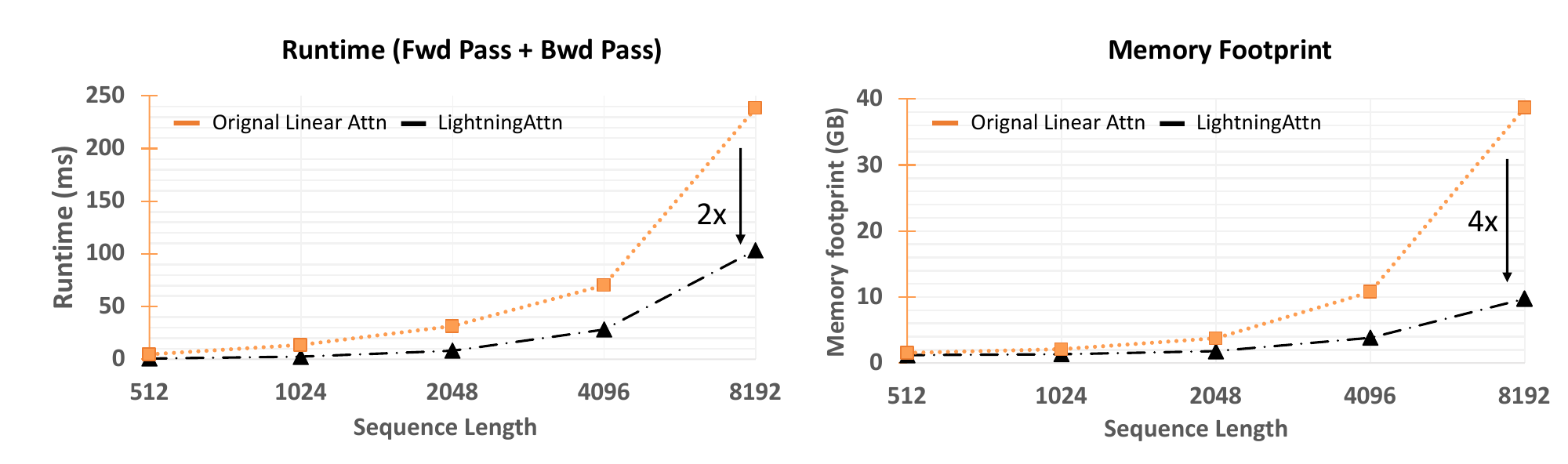}
\vspace{-4mm}
\caption{\textbf{Memory and speed comparison between linear attention and lightning attention.} Left: runtime of forward + backward pass milliseconds for different sequence lengths, with a fixed feature dimension of 2048. Right: memory footprints of forward + backward pass for different sequence lengths, with a fixed feature dimension of 2048.}
\vspace{-4mm}
\label{fig: flash}
\end{figure}

\vspace{-2mm}
\paragraph{Lightning Attention} We conducted a speed and memory comparison between our Lightning Attention and the baseline, which is the PyTorch implementation of the NormAttention~\citep{qin-etal-2022-devil}.
Figure~\ref{fig: flash} (left) reports the runtime in milliseconds of the forward + backward pass.
Baseline runtime grows quadratically with sequence length, while Lightning Attention operates significantly faster, at least $2\times$ faster than the PyTorch implementation. 
Figure~\ref{fig: flash} (right) reports the memory footprint of Lightning Attention compared to the baseline.
The memory footprint of Lightning Attention grows linearly with sequence length, which is up to $4\times$ more efficient than the baseline when the sequence length is 8192.
Our proposed Lightning Attention achieves superior efficiency.

\begin{figure}[H]
    \centering
    \includegraphics[width=0.9\textwidth]{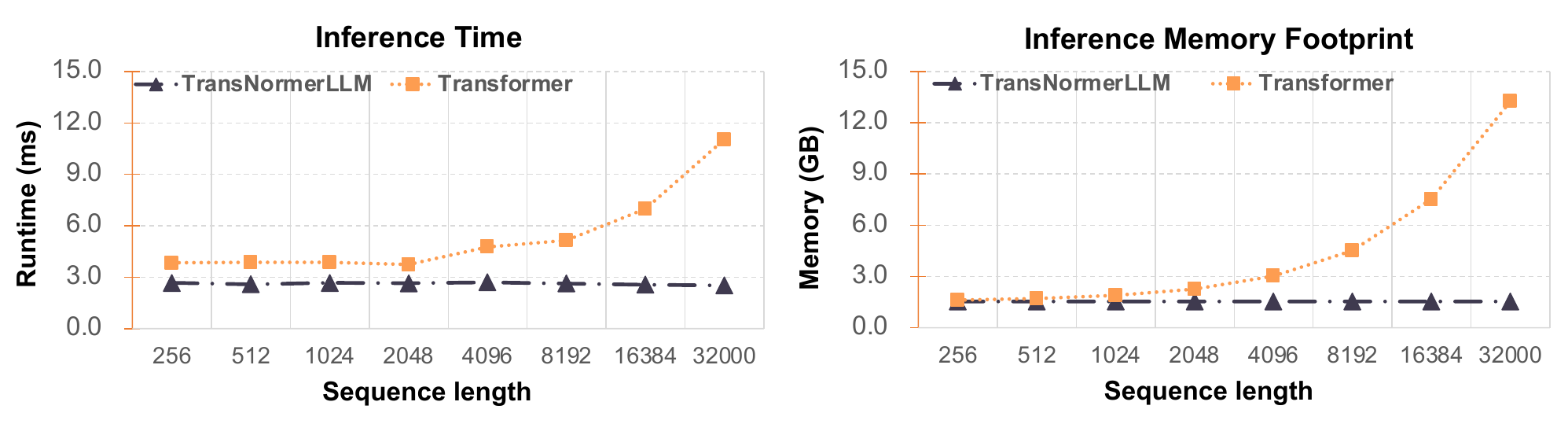}
    \vspace{-4mm}
     \caption{\textbf{Inference Time and Memory Footprint.} Left: inference runtime measured in milliseconds
     across different sequence lengths. Right: memory consumption during inference for varying sequence lengths. It is noteworthy that as the sequence length increases, TransNormerLLM demonstrates a consistent inference time and memory footprint.}
    \vspace{-4mm}
\end{figure}

\subsection{Benchmarks}
\vspace{-2mm}

\definecolor{Gray}{gray}{0.9}
\begin{table}[!ht]
    \centering
    \vspace{-8mm}
    \caption{\textbf{Performance Comparison on Commonsense Reasoning and Aggregated Benchmarks.} For a fair comparison, we report competing methods' results reproduced by us using their released models. 
    Official results are denoted in \textit{italics}. PS: parameter size (billion). T: tokens (trillion).
    HS: HellaSwag. WG: WinoGrande. }
    \scalebox{0.75}{
   
    \begin{tabular}{p{2cm}|cc|cccccccccc}
    \hline
        Model & PS & T & BoolQ & PIQA & HS & WG & ARC-e & ARC-c & OBQA & MMLU & CMMLU & C-Eval \\ \hline
        OPT & 0.35 & 0.30 & 57.74 & 64.58 & 36.69 & 52.49 & 44.02 & 23.89 & 28.20 & 26.02 & 25.34 & 25.71 \\ 
        Pythia & 0.40 & 0.30 & 60.40 & 67.08 & 40.52 & 53.59 & 51.81 & 24.15 & 29.40 & 25.99 & 25.16 & 24.81 \\ 
        BLOOM & 0.56 & 0.35 & 55.14 & 64.09 & 36.97 & 52.80 & 47.35 & 23.98 & 28.20 & 24.80 & 25.35 & 27.14 \\ 
        RWKV & 0.43 & - & - & \textit{67.52} & \textit{40.90} & \textit{51.14} & \textit{52.86} & \textit{25.17} & \textit{32.40} & 24.85 & - & - \\
               \rowcolor{Gray}
        Ours & 0.39 & 1.0 &62.14 &	66.70	 &46.27	 &54.46	 &55.43 &	27.99 &32.40 &25.90 & 25.05 & 25.24 \\ 
        \hline

        GPT-Neo & 1.3 & 0.3 & 61.99 & 71.11 & 48.93 & 54.93 & 56.19 & 25.85 & 33.60 & 24.82 & 26.03 & 23.94 \\ 
        OPT & 1.3 & 0.3 & 57.77 & 71.71 & 53.70 & 59.35 & 57.24 & 29.69 & 33.20 & 24.96 & 24.97 & 25.32 \\ 
        Pythia & 1.4 & 0.3 & 60.73 & 70.67 & 47.18 & 53.51 & 56.99 & 26.88 & 31.40 & 26.55 & 25.13 & 24.25 \\
        BLOOM & 1.1 & 0.35 & 59.08 & 67.14 & 42.98 & 54.93 & 51.47 & 25.68 & 29.40 & 27.30 & 25.09 & 26.50 \\
        RWKV & 1.5 &- & - & \textit{72.36} & \textit{52.48} & \textit{54.62} & \textit{60.48} & \textit{29.44} & \textit{34.00} & 25.77 & - & - \\ 
        Falcon & 1.0 & 0.35 & 61.38 & 75.14 & 61.50 & 60.30 & 63.38 & 32.17 & 35.60 & 25.28 & 24.88 & 25.66 \\ 
             \rowcolor{Gray}
        Ours & 1.0 & 1.2 & 63.27& 72.09& 56.49& 60.38& 63.68& 35.24& 36.60 & 27.10 & 25.88 &26.01\\
        
        \hline

        GPT-J & 6.9 & 0.3 & 65.44& 75.41& 66.25&64.09&66.92&36.60&38.20&25.40&26.47&23.39\\ 
        OPT & 6.7 & 0.3 & 66.18 & 76.22 & 67.21 & 65.19 & 65.66 & 34.64 & 37.20 & 24.57 & 25.36 & 25.32 \\ 
        Pythia & 6.9 & 0.3 & 63.46 & 75.14 & 63.92 & 60.77 & 67.34 & 35.41 & 37.00 & 24.64 & 25.56 & 26.40 \\ 
        BLOOM & 7.1 & 0.35 & 62.91 & 72.69 & 62.33 & 64.01 & 65.11 & 33.45 & 35.80 & 26.25 & 24.97 & 24.25 \\ 
        RWKV & 7.4 & - & - & \textit{76.06} & \textit{65.51} & \textit{61.01} & \textit{67.80} & \textit{37.46} & \textit{40.20} & 24.96 & - & - \\ 
        MPT & 6.9 & 1.0 & 73.88 & 79.43 & 76.25 & 68.27 & 74.79 & 41.72 & 42.20 & 30.80 & 25.99 & 24.06 \\ 
        Falcon & 7.2 & 1.5 & 73.73 & 79.38 & 76.3 & 67.17 & 74.62 & 43.60 & 43.80 & 27.79 & 25.73 & 22.92 \\ 
    
        Baichuan1 & 7.0 & 1.2 & 70.09 & 76.01 & 70.06 & 64.09 & 71.72 & 40.53 & 38.20 & \textit{42.30} & \textit{44.43} & \textit{42.80} \\ 
        Baichuan2 & 7.0 & 2.6 & 72.72 & 76.50 & 72.17 & 68.35 & 75.17 & 42.32 & 39.60 & \textit{54.16} & \textit{57.07} & \textit{54.00} \\ 
        ChatGLM1 & 6.7 & 1.0 & 74.74 & 68.88 & 45.57 & 52.25 & 48.78 & 31.66 & 36.80 & \textit{40.63} & 37.48 & \textit{40.23} \\ 
        ChatGLM2 & 7.1 & 1.4 & 77.65 & 69.37 & 50.51 & 57.62 & 59.13 & 34.30 & 37.00 & \textit{45.46} & 48.80 & \textit{52.55} \\
       
        OpenLLaMAv1 & 6.7 & 1.0 & 70.43 & 75.68 & 69.23 & 66.69 & 71.17 & 38.57 & 39.00 & 30.49 & 25.40 & 26.09 \\ 
       OpenLLaMAv2 & 6.7 & 1.0 & 72.20 & 78.84 & 74.51 & 65.67 & 72.39 & 41.30 & 41.00 & 41.29 & 29.58 & 30.01 \\ 
    LLaMA1 & 6.7 & 1.0 & \textit{76.50} & \textit{79.80} & \textit{76.10} & \textit{70.10} & \textit{72.80} & \textit{47.60} & \textit{57.20} & \textit{35.10} & 25.62 & 25.72 \\ 
        LLaMA2 & 6.7 & 2.0 & \textit{77.68} & \textit{78.07} & \textit{76.02} & \textit{68.98} & \textit{76.30} & \textit{46.33} & \textit{44.20} & \textit{45.30} & 32.96 & 33.20 \\ 
         \rowcolor{Gray}
       Ours & 6.8 & 1.4 &75.87&	80.09 &75.21 & 66.06 &75.42 & 44.40 & 63.40 & 43.10 & 47.99 & 43.18 \\
        \hline
    \end{tabular}
}
\vspace{-4mm}
\end{table}

In order to validate the effectiveness of TransNormerLLM, we tested our 385M, 1B, and 7B models on Commonsense Reasoning Task, MMLU\citep{hendrycks2021measuring}, CMMLU\citep{li2023cmmlu}, and C-Eval\citep{huang2023ceval}. For comparison, we selected several open-source models as competitors, including Transformer-based models such as OPT~\citep{zhang2022opt}, Pythia~\citep{biderman2023pythia}, BLOOM~\citep{workshop2023bloom}, GPT-Neo~\citep{gpt-neo}, GPT-J~\citep{wang2021gpt}, MPT~\citep{mpt-7b}, Falcon~\citep{falcon40b}, LLaMA1/2~\citep{touvron2023llama,2307.09288}, OpenLLAMA v1/v2~\citep{openlm2023openllama}, Baichuan 1/2~\citep{baichuan2023baichuan2}, ChatGLM 1/2~\citep{zeng2022glm,du2022glm}, and non-Transformer model RWKV~\citep{2305.13048}. It can be observed that, compared to these models, TransNormerLLM remains highly competitive.

\vspace{-3mm}
\paragraph{Commonsense Reasoning} We report BoolQ \citep{clark2019boolq}, PIQA \citep{bisk2019piqa}, SIQA \citep{sap2019socialiqa},
HellaSwag \citep{zellers2019hellaswag}, WinoGrande \citep{sakaguchi2019winogrande}, ARC easy and challenge \citep{clark2018think}, OpenBookQA \citep{mihaylov2018suit} and their average. We report 0-shot results for all benchmarks using LM-Eval-Harness \citep{leo2021evalharness}.
All of our models achieve competitive performance compared to existing state-of-the-art LLMs, showcasing a remarkable ability to comprehend and apply commonsense reasoning.

\vspace{-3mm}
\paragraph{Aggregated Benchmarks}
 We report the overall results for MMLU \citep{hendrycks2021measuring}, CMMLU \citep{li2023cmmlu}, C-Eval \citep{huang2023ceval}. Official scripts were used for evaluating MMLU, CMMLU, and C-Eval, with all evaluation results being conducted with a 5-shot setup. In comparison to top-tier open-source models available in the industry, our models have demonstrated matched performance in both English and Chinese benchmarks.

\vspace{-2mm}
\subsection{Scaling to 175B}
\vspace{-2mm}
Furthermore, we have carried out a series of experiments to assess the efficacy of model parallelism as applied to the TransNormerLLM architecture. The comprehensive outcomes of these experiments have been thoughtfully presented in Appendix~\ref{app:model_para}.
Moreover, our research extends to the meticulous evaluation of various cutting-edge system optimization techniques. This evaluation encompasses their impact on both training speed and context length across models ranging from 7B to 175B in scale. We have thoughtfully documented the detailed results of these experiments in Appendix~\ref{app:size_length}.
\vspace{-3mm}

\section{Conclusion}
\vspace{-2mm}
We introduced TransNormerLLM in this paper, an improved TransNormer that is tailored for LLMs. Our TransNormerLLM consistently outperformed Transformers in both accuracy and efficiency. Extensive ablations demonstrate the effectiveness of our modifications and innovations in position encoding, gating mechanism, activation functions, normalization functions, and lightning attentions. 
These modifications collectively contribute to TransNormerLLM's outstanding performance, positioning it as a promising choice for state-of-the-art language models. 
The benchmark results for models with sizes of 385 million, 1 billion, and 7 billion parameters unequivocally demonstrate that TransNormerLLM not only matches the performance of current leading Transformer-based Large Language Models (LLMs) but also enjoys faster inference speeds.
We will release our pre-trained TransNormerLLM models to foster community advancements in efficient LLM. 

\bibliography{iclr2024_conference}

\begin{thebibliography}{71}
\providecommand{\natexlab}[1]{#1}
\providecommand{\url}[1]{\texttt{#1}}
\expandafter\ifx\csname urlstyle\endcsname\relax
  \providecommand{\doi}[1]{doi: #1}\else
  \providecommand{\doi}{doi: \begingroup \urlstyle{rm}\Url}\fi

\bibitem[Almazrouei et~al.(2023)Almazrouei, Alobeidli, Alshamsi, Cappelli, Cojocaru, Debbah, Goffinet, Heslow, Launay, Malartic, et~al.]{falcon40b}
Ebtesam Almazrouei, Hamza Alobeidli, Abdulaziz Alshamsi, Alessandro Cappelli, Ruxandra Cojocaru, Merouane Debbah, Etienne Goffinet, Daniel Heslow, Julien Launay, Quentin Malartic, et~al.
\newblock Falcon-40b: an open large language model with state-of-the-art performance.
\newblock Technical report, Technical report, Technology Innovation Institute, 2023.

\bibitem[Baichuan(2023)]{baichuan2023baichuan2}
Baichuan.
\newblock Baichuan 2: Open large-scale language models.
\newblock \emph{arXiv preprint arXiv:2309.10305}, 2023.
\newblock URL \url{https://arxiv.org/abs/2309.10305}.

\bibitem[Beltagy et~al.(2020)Beltagy, Peters, and Cohan]{beltagy2020longformer}
Iz~Beltagy, Matthew~E. Peters, and Arman Cohan.
\newblock Longformer: The long-document transformer, 2020.

\bibitem[Biderman et~al.(2023)Biderman, Schoelkopf, Anthony, Bradley, O'Brien, Hallahan, Khan, Purohit, Prashanth, Raff, Skowron, Sutawika, and van~der Wal]{biderman2023pythia}
Stella Biderman, Hailey Schoelkopf, Quentin Anthony, Herbie Bradley, Kyle O'Brien, Eric Hallahan, Mohammad~Aflah Khan, Shivanshu Purohit, USVSN~Sai Prashanth, Edward Raff, Aviya Skowron, Lintang Sutawika, and Oskar van~der Wal.
\newblock Pythia: A suite for analyzing large language models across training and scaling, 2023.

\bibitem[Bisk et~al.(2019)Bisk, Zellers, Bras, Gao, and Choi]{bisk2019piqa}
Yonatan Bisk, Rowan Zellers, Ronan~Le Bras, Jianfeng Gao, and Yejin Choi.
\newblock Piqa: Reasoning about physical commonsense in natural language, 2019.

\bibitem[Black et~al.(2022)Black, Biderman, Hallahan, Anthony, Gao, Golding, He, Leahy, McDonell, Phang, et~al.]{gpt-neo}
Sid Black, Stella Biderman, Eric Hallahan, Quentin Anthony, Leo Gao, Laurence Golding, Horace He, Connor Leahy, Kyle McDonell, Jason Phang, et~al.
\newblock Gpt-neox-20b: An open-source autoregressive language model.
\newblock \emph{arXiv preprint arXiv:2204.06745}, 2022.

\bibitem[Brown et~al.(2020)Brown, Mann, Ryder, Subbiah, Kaplan, Dhariwal, Neelakantan, Shyam, Sastry, Askell, et~al.]{brown2020language}
Tom Brown, Benjamin Mann, Nick Ryder, Melanie Subbiah, Jared~D Kaplan, Prafulla Dhariwal, Arvind Neelakantan, Pranav Shyam, Girish Sastry, Amanda Askell, et~al.
\newblock Language models are few-shot learners.
\newblock \emph{Advances in neural information processing systems}, 33:\penalty0 1877--1901, 2020.

\bibitem[Child et~al.(2019)Child, Gray, Radford, and Sutskever]{1904.10509}
Rewon Child, Scott Gray, Alec Radford, and Ilya Sutskever.
\newblock Generating long sequences with sparse transformers, 2019.

\bibitem[Choromanski et~al.(2021)Choromanski, Likhosherstov, Dohan, Song, Gane, Sarlos, Hawkins, Davis, Mohiuddin, Kaiser, Belanger, Colwell, and Weller]{choromanski2021rethinking}
Krzysztof~Marcin Choromanski, Valerii Likhosherstov, David Dohan, Xingyou Song, Andreea Gane, Tamas Sarlos, Peter Hawkins, Jared~Quincy Davis, Afroz Mohiuddin, Lukasz Kaiser, David~Benjamin Belanger, Lucy~J Colwell, and Adrian Weller.
\newblock Rethinking attention with performers.
\newblock In \emph{International Conference on Learning Representations}, 2021.
\newblock URL \url{https://openreview.net/forum?id=Ua6zuk0WRH}.

\bibitem[Chowdhery et~al.(2022)Chowdhery, Narang, Devlin, Bosma, Mishra, Roberts, Barham, Chung, Sutton, Gehrmann, Schuh, Shi, Tsvyashchenko, Maynez, Rao, Barnes, Tay, Shazeer, Prabhakaran, Reif, Du, Hutchinson, Pope, Bradbury, Austin, Isard, Gur-Ari, Yin, Duke, Levskaya, Ghemawat, Dev, Michalewski, Garcia, Misra, Robinson, Fedus, Zhou, Ippolito, Luan, Lim, Zoph, Spiridonov, Sepassi, Dohan, Agrawal, Omernick, Dai, Pillai, Pellat, Lewkowycz, Moreira, Child, Polozov, Lee, Zhou, Wang, Saeta, Diaz, Firat, Catasta, Wei, Meier-Hellstern, Eck, Dean, Petrov, and Fiedel]{2204.02311}
Aakanksha Chowdhery, Sharan Narang, Jacob Devlin, Maarten Bosma, Gaurav Mishra, Adam Roberts, Paul Barham, Hyung~Won Chung, Charles Sutton, Sebastian Gehrmann, Parker Schuh, Kensen Shi, Sasha Tsvyashchenko, Joshua Maynez, Abhishek Rao, Parker Barnes, Yi~Tay, Noam Shazeer, Vinodkumar Prabhakaran, Emily Reif, Nan Du, Ben Hutchinson, Reiner Pope, James Bradbury, Jacob Austin, Michael Isard, Guy Gur-Ari, Pengcheng Yin, Toju Duke, Anselm Levskaya, Sanjay Ghemawat, Sunipa Dev, Henryk Michalewski, Xavier Garcia, Vedant Misra, Kevin Robinson, Liam Fedus, Denny Zhou, Daphne Ippolito, David Luan, Hyeontaek Lim, Barret Zoph, Alexander Spiridonov, Ryan Sepassi, David Dohan, Shivani Agrawal, Mark Omernick, Andrew~M. Dai, Thanumalayan~Sankaranarayana Pillai, Marie Pellat, Aitor Lewkowycz, Erica Moreira, Rewon Child, Oleksandr Polozov, Katherine Lee, Zongwei Zhou, Xuezhi Wang, Brennan Saeta, Mark Diaz, Orhan Firat, Michele Catasta, Jason Wei, Kathy Meier-Hellstern, Douglas Eck, Jeff Dean, Slav Petrov, and Noah Fiedel.
\newblock Palm: Scaling language modeling with pathways, 2022.

\bibitem[Clark et~al.(2019)Clark, Lee, Chang, Kwiatkowski, Collins, and Toutanova]{clark2019boolq}
Christopher Clark, Kenton Lee, Ming-Wei Chang, Tom Kwiatkowski, Michael Collins, and Kristina Toutanova.
\newblock Boolq: Exploring the surprising difficulty of natural yes/no questions, 2019.

\bibitem[Clark et~al.(2018)Clark, Cowhey, Etzioni, Khot, Sabharwal, Schoenick, and Tafjord]{clark2018think}
Peter Clark, Isaac Cowhey, Oren Etzioni, Tushar Khot, Ashish Sabharwal, Carissa Schoenick, and Oyvind Tafjord.
\newblock Think you have solved question answering? try arc, the ai2 reasoning challenge, 2018.

\bibitem[Dao(2023)]{dao2023flashattention2}
Tri Dao.
\newblock Flashattention-2: Faster attention with better parallelism and work partitioning.
\newblock \emph{arXiv preprint arXiv:2307.08691}, 2023.

\bibitem[Dao et~al.(2022{\natexlab{a}})Dao, Fu, Ermon, Rudra, and R{\'e}]{dao2022flashattention}
Tri Dao, Daniel~Y. Fu, Stefano Ermon, Atri Rudra, and Christopher R{\'e}.
\newblock Flash{A}ttention: Fast and memory-efficient exact attention with {IO}-awareness.
\newblock In \emph{Advances in Neural Information Processing Systems}, 2022{\natexlab{a}}.

\bibitem[Dao et~al.(2022{\natexlab{b}})Dao, Fu, Saab, Thomas, Rudra, and R{\'{e}}]{h3}
Tri Dao, Daniel~Y. Fu, Khaled~Kamal Saab, Armin~W. Thomas, Atri Rudra, and Christopher R{\'{e}}.
\newblock Hungry hungry hippos: Towards language modeling with state space models.
\newblock \emph{CoRR}, abs/2212.14052, 2022{\natexlab{b}}.
\newblock \doi{10.48550/arXiv.2212.14052}.
\newblock URL \url{https://doi.org/10.48550/arXiv.2212.14052}.

\bibitem[Devlin et~al.(2019)Devlin, Chang, Lee, and Toutanova]{devlin-etal-2019-bert}
Jacob Devlin, Ming-Wei Chang, Kenton Lee, and Kristina Toutanova.
\newblock {BERT}: Pre-training of deep bidirectional transformers for language understanding.
\newblock In \emph{Proceedings of the 2019 Conference of the North {A}merican Chapter of the Association for Computational Linguistics: Human Language Technologies, Volume 1 (Long and Short Papers)}, pp.\  4171--4186, Minneapolis, Minnesota, June 2019. Association for Computational Linguistics.
\newblock \doi{10.18653/v1/N19-1423}.
\newblock URL \url{https://aclanthology.org/N19-1423}.

\bibitem[Du et~al.(2022)Du, Qian, Liu, Ding, Qiu, Yang, and Tang]{du2022glm}
Zhengxiao Du, Yujie Qian, Xiao Liu, Ming Ding, Jiezhong Qiu, Zhilin Yang, and Jie Tang.
\newblock Glm: General language model pretraining with autoregressive blank infilling, 2022.

\bibitem[Fu et~al.(2023)Fu, Epstein, Nguyen, Thomas, Zhang, Dao, Rudra, and R{\'{e}}]{simplelongconv}
Daniel~Y. Fu, Elliot~L. Epstein, Eric Nguyen, Armin~W. Thomas, Michael Zhang, Tri Dao, Atri Rudra, and Christopher R{\'{e}}.
\newblock Simple hardware-efficient long convolutions for sequence modeling.
\newblock \emph{CoRR}, abs/2302.06646, 2023.
\newblock \doi{10.48550/arXiv.2302.06646}.
\newblock URL \url{https://doi.org/10.48550/arXiv.2302.06646}.

\bibitem[Gao et~al.(2021)Gao, Tow, Biderman, Black, DiPofi, Foster, Golding, Hsu, McDonell, Muennighoff, et~al.]{leo2021evalharness}
Leo Gao, Jonathan Tow, Stella Biderman, Sid Black, Anthony DiPofi, Charles Foster, Laurence Golding, Jeffrey Hsu, Kyle McDonell, Niklas Muennighoff, et~al.
\newblock A framework for few-shot language model evaluation.
\newblock \emph{Version v0. 0.1. Sept}, 2021.

\bibitem[Geng \& Liu(2023)Geng and Liu]{openlm2023openllama}
Xinyang Geng and Hao Liu.
\newblock Openllama: An open reproduction of llama.
\newblock \emph{URL: https://github. com/openlm-research/open\_llama}, 2023.

\bibitem[Gu et~al.(2020)Gu, Dao, Ermon, Rudra, and Re]{2008.07669}
Albert Gu, Tri Dao, Stefano Ermon, Atri Rudra, and Christopher Re.
\newblock Hippo: Recurrent memory with optimal polynomial projections, 2020.

\bibitem[Gu et~al.(2022{\natexlab{a}})Gu, Goel, Gupta, and R{\'{e}}]{s4d}
Albert Gu, Karan Goel, Ankit Gupta, and Christopher R{\'{e}}.
\newblock On the parameterization and initialization of diagonal state space models.
\newblock In \emph{NeurIPS}, 2022{\natexlab{a}}.
\newblock URL \url{http://papers.nips.cc/paper\_files/paper/2022/hash/e9a32fade47b906de908431991440f7c-Abstract-Conference.html}.

\bibitem[Gu et~al.(2022{\natexlab{b}})Gu, Goel, and R{\'{e}}]{s4}
Albert Gu, Karan Goel, and Christopher R{\'{e}}.
\newblock Efficiently modeling long sequences with structured state spaces.
\newblock In \emph{The Tenth International Conference on Learning Representations, {ICLR} 2022, Virtual Event, April 25-29, 2022}. OpenReview.net, 2022{\natexlab{b}}.
\newblock URL \url{https://openreview.net/forum?id=uYLFoz1vlAC}.

\bibitem[Gupta et~al.(2022)Gupta, Gu, and Berant]{gupta2022DSS}
Ankit Gupta, Albert Gu, and Jonathan Berant.
\newblock Diagonal state spaces are as effective as structured state spaces.
\newblock In \emph{NeurIPS}, 2022.
\newblock URL \url{http://papers.nips.cc/paper\_files/paper/2022/hash/9156b0f6dfa9bbd18c79cc459ef5d61c-Abstract-Conference.html}.

\bibitem[Hendrycks et~al.(2021)Hendrycks, Burns, Basart, Zou, Mazeika, Song, and Steinhardt]{hendrycks2021measuring}
Dan Hendrycks, Collin Burns, Steven Basart, Andy Zou, Mantas Mazeika, Dawn Song, and Jacob Steinhardt.
\newblock Measuring massive multitask language understanding, 2021.

\bibitem[Hoffmann et~al.(2022)Hoffmann, Borgeaud, Mensch, Buchatskaya, Cai, Rutherford, de~Las~Casas, Hendricks, Welbl, Clark, Hennigan, Noland, Millican, van~den Driessche, Damoc, Guy, Osindero, Simonyan, Elsen, Rae, Vinyals, and Sifre]{hoffmann2022training}
Jordan Hoffmann, Sebastian Borgeaud, Arthur Mensch, Elena Buchatskaya, Trevor Cai, Eliza Rutherford, Diego de~Las~Casas, Lisa~Anne Hendricks, Johannes Welbl, Aidan Clark, Tom Hennigan, Eric Noland, Katie Millican, George van~den Driessche, Bogdan Damoc, Aurelia Guy, Simon Osindero, Karen Simonyan, Erich Elsen, Jack~W. Rae, Oriol Vinyals, and Laurent Sifre.
\newblock Training compute-optimal large language models, 2022.

\bibitem[Hua et~al.(2022)Hua, Dai, Liu, and Le]{hua2022transformer}
Weizhe Hua, Zihang Dai, Hanxiao Liu, and Quoc~V Le.
\newblock Transformer quality in linear time.
\newblock \emph{arXiv preprint arXiv:2202.10447}, 2022.

\bibitem[Huang et~al.(2023)Huang, Bai, Zhu, Zhang, Zhang, Su, Liu, Lv, Zhang, Lei, Fu, Sun, and He]{huang2023ceval}
Yuzhen Huang, Yuzhuo Bai, Zhihao Zhu, Junlei Zhang, Jinghan Zhang, Tangjun Su, Junteng Liu, Chuancheng Lv, Yikai Zhang, Jiayi Lei, Yao Fu, Maosong Sun, and Junxian He.
\newblock C-eval: A multi-level multi-discipline chinese evaluation suite for foundation models, 2023.

\bibitem[Kalamkar et~al.(2019)Kalamkar, Mudigere, Mellempudi, Das, Banerjee, Avancha, Vooturi, Jammalamadaka, Huang, Yuen, et~al.]{kalamkar2019study}
Dhiraj Kalamkar, Dheevatsa Mudigere, Naveen Mellempudi, Dipankar Das, Kunal Banerjee, Sasikanth Avancha, Dharma~Teja Vooturi, Nataraj Jammalamadaka, Jianyu Huang, Hector Yuen, et~al.
\newblock A study of bfloat16 for deep learning training.
\newblock \emph{arXiv preprint arXiv:1905.12322}, 2019.

\bibitem[Kaplan et~al.(2020)Kaplan, McCandlish, Henighan, Brown, Chess, Child, Gray, Radford, Wu, and Amodei]{kaplan2020scaling}
Jared Kaplan, Sam McCandlish, Tom Henighan, Tom~B. Brown, Benjamin Chess, Rewon Child, Scott Gray, Alec Radford, Jeffrey Wu, and Dario Amodei.
\newblock Scaling laws for neural language models, 2020.

\bibitem[Katharopoulos et~al.(2020)Katharopoulos, Vyas, Pappas, and Fleuret]{katharopoulos2020transformers}
Angelos Katharopoulos, Apoorv Vyas, Nikolaos Pappas, and Fran{\c{c}}ois Fleuret.
\newblock Transformers are rnns: Fast autoregressive transformers with linear attention.
\newblock In \emph{International Conference on Machine Learning}, pp.\  5156--5165. PMLR, 2020.

\bibitem[Ke et~al.(2021)Ke, He, and Liu]{ke2021rethinking}
Guolin Ke, Di~He, and Tie-Yan Liu.
\newblock Rethinking positional encoding in language pre-training.
\newblock In \emph{International Conference on Learning Representations}, 2021.
\newblock URL \url{https://openreview.net/forum?id=09-528y2Fgf}.

\bibitem[Kingma \& Ba(2017)Kingma and Ba]{kingma2017adam}
Diederik~P. Kingma and Jimmy Ba.
\newblock Adam: A method for stochastic optimization, 2017.

\bibitem[Lewis et~al.(2019)Lewis, Liu, Goyal, Ghazvininejad, Mohamed, Levy, Stoyanov, and Zettlemoyer]{lewis2019bart}
Mike Lewis, Yinhan Liu, Naman Goyal, Marjan Ghazvininejad, Abdelrahman Mohamed, Omer Levy, Ves Stoyanov, and Luke Zettlemoyer.
\newblock Bart: Denoising sequence-to-sequence pre-training for natural language generation, translation, and comprehension, 2019.

\bibitem[Li et~al.(2023)Li, Zhang, Koto, Yang, Zhao, Gong, Duan, and Baldwin]{li2023cmmlu}
Haonan Li, Yixuan Zhang, Fajri Koto, Yifei Yang, Hai Zhao, Yeyun Gong, Nan Duan, and Timothy Baldwin.
\newblock Cmmlu: Measuring massive multitask language understanding in chinese, 2023.

\bibitem[Liu et~al.(2022)Liu, Li, Lu, Qin, Sun, Xu, and Zhong]{liu2022neural}
Zexiang Liu, Dong Li, Kaiyue Lu, Zhen Qin, Weixuan Sun, Jiacheng Xu, and Yiran Zhong.
\newblock Neural architecture search on efficient transformers and beyond.
\newblock \emph{arXiv preprint arXiv:2207.13955}, 2022.

\bibitem[Micikevicius et~al.(2017)Micikevicius, Narang, Alben, Diamos, Elsen, Garcia, Ginsburg, Houston, Kuchaiev, Venkatesh, et~al.]{micikevicius2017mixed}
Paulius Micikevicius, Sharan Narang, Jonah Alben, Gregory Diamos, Erich Elsen, David Garcia, Boris Ginsburg, Michael Houston, Oleksii Kuchaiev, Ganesh Venkatesh, et~al.
\newblock Mixed precision training.
\newblock \emph{arXiv preprint arXiv:1710.03740}, 2017.

\bibitem[Mihaylov et~al.(2018)Mihaylov, Clark, Khot, and Sabharwal]{mihaylov2018suit}
Todor Mihaylov, Peter Clark, Tushar Khot, and Ashish Sabharwal.
\newblock Can a suit of armor conduct electricity? a new dataset for open book question answering, 2018.

\bibitem[Orvieto et~al.(2023)Orvieto, Smith, Gu, Fernando, Gulcehre, Pascanu, and De]{2303.06349}
Antonio Orvieto, Samuel~L Smith, Albert Gu, Anushan Fernando, Caglar Gulcehre, Razvan Pascanu, and Soham De.
\newblock Resurrecting recurrent neural networks for long sequences, 2023.

\bibitem[Paszke et~al.(2019)Paszke, Gross, Massa, Lerer, Bradbury, Chanan, Killeen, Lin, Gimelshein, Antiga, Desmaison, Kopf, Yang, DeVito, Raison, Tejani, Chilamkurthy, Steiner, Fang, Bai, and Chintala]{NEURIPS2019_9015}
Adam Paszke, Sam Gross, Francisco Massa, Adam Lerer, James Bradbury, Gregory Chanan, Trevor Killeen, Zeming Lin, Natalia Gimelshein, Luca Antiga, Alban Desmaison, Andreas Kopf, Edward Yang, Zachary DeVito, Martin Raison, Alykhan Tejani, Sasank Chilamkurthy, Benoit Steiner, Lu~Fang, Junjie Bai, and Soumith Chintala.
\newblock Pytorch: An imperative style, high-performance deep learning library.
\newblock In \emph{Advances in Neural Information Processing Systems 32}, pp.\  8024--8035. Curran Associates, Inc., 2019.
\newblock URL \url{http://papers.neurips.cc/paper/9015-pytorch-an-imperative-style-high-performance-deep-learning-library.pdf}.

\bibitem[Penedo et~al.(2023)Penedo, Malartic, Hesslow, Cojocaru, Cappelli, Alobeidli, Pannier, Almazrouei, and Launay]{penedo2023refinedweb}
Guilherme Penedo, Quentin Malartic, Daniel Hesslow, Ruxandra Cojocaru, Alessandro Cappelli, Hamza Alobeidli, Baptiste Pannier, Ebtesam Almazrouei, and Julien Launay.
\newblock The refinedweb dataset for falcon llm: Outperforming curated corpora with web data, and web data only, 2023.

\bibitem[Peng et~al.(2023{\natexlab{a}})Peng, Alcaide, Anthony, Albalak, Arcadinho, Cao, Cheng, Chung, Grella, GV, He, Hou, Kazienko, Kocon, Kong, Koptyra, Lau, Mantri, Mom, Saito, Tang, Wang, Wind, Wozniak, Zhang, Zhang, Zhao, Zhou, Zhu, and Zhu]{2305.13048}
Bo~Peng, Eric Alcaide, Quentin Anthony, Alon Albalak, Samuel Arcadinho, Huanqi Cao, Xin Cheng, Michael Chung, Matteo Grella, Kranthi~Kiran GV, Xuzheng He, Haowen Hou, Przemyslaw Kazienko, Jan Kocon, Jiaming Kong, Bartlomiej Koptyra, Hayden Lau, Krishna Sri~Ipsit Mantri, Ferdinand Mom, Atsushi Saito, Xiangru Tang, Bolun Wang, Johan~S. Wind, Stansilaw Wozniak, Ruichong Zhang, Zhenyuan Zhang, Qihang Zhao, Peng Zhou, Jian Zhu, and Rui-Jie Zhu.
\newblock Rwkv: Reinventing rnns for the transformer era, 2023{\natexlab{a}}.

\bibitem[Peng et~al.(2023{\natexlab{b}})Peng, Alcaide, Anthony, Albalak, Arcadinho, Cao, Cheng, Chung, Grella, GV, He, Hou, Kazienko, Kocon, Kong, Koptyra, Lau, Mantri, Mom, Saito, Tang, Wang, Wind, Wozniak, Zhang, Zhang, Zhao, Zhou, Zhu, and Zhu]{peng2023rwkv}
Bo~Peng, Eric Alcaide, Quentin Anthony, Alon Albalak, Samuel Arcadinho, Huanqi Cao, Xin Cheng, Michael Chung, Matteo Grella, Kranthi~Kiran GV, Xuzheng He, Haowen Hou, Przemyslaw Kazienko, Jan Kocon, Jiaming Kong, Bartlomiej Koptyra, Hayden Lau, Krishna Sri~Ipsit Mantri, Ferdinand Mom, Atsushi Saito, Xiangru Tang, Bolun Wang, Johan~S. Wind, Stansilaw Wozniak, Ruichong Zhang, Zhenyuan Zhang, Qihang Zhao, Peng Zhou, Jian Zhu, and Rui-Jie Zhu.
\newblock Rwkv: Reinventing rnns for the transformer era, 2023{\natexlab{b}}.

\bibitem[Press et~al.(2022)Press, Smith, and Lewis]{alibi}
Ofir Press, Noah Smith, and Mike Lewis.
\newblock Train short, test long: Attention with linear biases enables input length extrapolation.
\newblock In \emph{International Conference on Learning Representations}, 2022.
\newblock URL \url{https://openreview.net/forum?id=R8sQPpGCv0}.

\bibitem[Qin et~al.(2022{\natexlab{a}})Qin, Han, Sun, Li, Kong, Barnes, and Zhong]{qin-etal-2022-devil}
Zhen Qin, Xiaodong Han, Weixuan Sun, Dongxu Li, Lingpeng Kong, Nick Barnes, and Yiran Zhong.
\newblock The devil in linear transformer.
\newblock In \emph{Proceedings of the 2022 Conference on Empirical Methods in Natural Language Processing}, pp.\  7025--7041, Abu Dhabi, United Arab Emirates, December 2022{\natexlab{a}}. Association for Computational Linguistics.
\newblock URL \url{https://aclanthology.org/2022.emnlp-main.473}.

\bibitem[Qin et~al.(2022{\natexlab{b}})Qin, Sun, Deng, Li, Wei, Lv, Yan, Kong, and Zhong]{zhen2022cosformer}
Zhen Qin, Weixuan Sun, Hui Deng, Dongxu Li, Yunshen Wei, Baohong Lv, Junjie Yan, Lingpeng Kong, and Yiran Zhong.
\newblock cosformer: Rethinking softmax in attention.
\newblock In \emph{International Conference on Learning Representations}, 2022{\natexlab{b}}.
\newblock URL \url{https://openreview.net/forum?id=Bl8CQrx2Up4}.

\bibitem[Qin et~al.(2023{\natexlab{a}})Qin, Han, Sun, He, Li, Li, Dai, Kong, and Zhong]{qin2023toeplitz}
Zhen Qin, Xiaodong Han, Weixuan Sun, Bowen He, Dong Li, Dongxu Li, Yuchao Dai, Lingpeng Kong, and Yiran Zhong.
\newblock Toeplitz neural network for sequence modeling.
\newblock In \emph{The Eleventh International Conference on Learning Representations}, 2023{\natexlab{a}}.
\newblock URL \url{https://openreview.net/forum?id=IxmWsm4xrua}.

\bibitem[Qin et~al.(2023{\natexlab{b}})Qin, Sun, Lu, Deng, Li, Han, Dai, Kong, and Zhong]{qin2023linearized}
Zhen Qin, Weixuan Sun, Kaiyue Lu, Hui Deng, Dongxu Li, Xiaodong Han, Yuchao Dai, Lingpeng Kong, and Yiran Zhong.
\newblock Linearized relative positional encoding.
\newblock \emph{Transactions on Machine Learning Research}, 2023{\natexlab{b}}.

\bibitem[Qin et~al.(2023{\natexlab{c}})Qin, Zhong, and Deng]{2307.10156}
Zhen Qin, Yiran Zhong, and Hui Deng.
\newblock Exploring transformer extrapolation, 2023{\natexlab{c}}.

\bibitem[Radford et~al.(2018)Radford, Narasimhan, Salimans, and Sutskever]{radford2018improving}
Alec Radford, Karthik Narasimhan, Tim Salimans, and Ilya Sutskever.
\newblock Improving language understanding by generative pre-training.
\newblock \url{https://s3-us-west-2.amazonaws.com/openai-assets/research-covers/language-unsupervised/language_understanding_paper.pdf}, 2018.

\bibitem[Radford et~al.(2019)Radford, Wu, Child, Luan, Amodei, Sutskever, et~al.]{radford2019language}
Alec Radford, Jeffrey Wu, Rewon Child, David Luan, Dario Amodei, Ilya Sutskever, et~al.
\newblock Language models are unsupervised multitask learners.
\newblock \emph{OpenAI blog}, 1\penalty0 (8):\penalty0 9, 2019.

\bibitem[Rae et~al.(2022)Rae, Borgeaud, Cai, Millican, Hoffmann, Song, Aslanides, Henderson, Ring, Young, Rutherford, Hennigan, Menick, Cassirer, Powell, van~den Driessche, Hendricks, Rauh, Huang, Glaese, Welbl, Dathathri, Huang, Uesato, Mellor, Higgins, Creswell, McAleese, Wu, Elsen, Jayakumar, Buchatskaya, Budden, Sutherland, Simonyan, Paganini, Sifre, Martens, Li, Kuncoro, Nematzadeh, Gribovskaya, Donato, Lazaridou, Mensch, Lespiau, Tsimpoukelli, Grigorev, Fritz, Sottiaux, Pajarskas, Pohlen, Gong, Toyama, de~Masson~d'Autume, Li, Terzi, Mikulik, Babuschkin, Clark, de~Las~Casas, Guy, Jones, Bradbury, Johnson, Hechtman, Weidinger, Gabriel, Isaac, Lockhart, Osindero, Rimell, Dyer, Vinyals, Ayoub, Stanway, Bennett, Hassabis, Kavukcuoglu, and Irving]{rae2022scaling}
Jack~W. Rae, Sebastian Borgeaud, Trevor Cai, Katie Millican, Jordan Hoffmann, Francis Song, John Aslanides, Sarah Henderson, Roman Ring, Susannah Young, Eliza Rutherford, Tom Hennigan, Jacob Menick, Albin Cassirer, Richard Powell, George van~den Driessche, Lisa~Anne Hendricks, Maribeth Rauh, Po-Sen Huang, Amelia Glaese, Johannes Welbl, Sumanth Dathathri, Saffron Huang, Jonathan Uesato, John Mellor, Irina Higgins, Antonia Creswell, Nat McAleese, Amy Wu, Erich Elsen, Siddhant Jayakumar, Elena Buchatskaya, David Budden, Esme Sutherland, Karen Simonyan, Michela Paganini, Laurent Sifre, Lena Martens, Xiang~Lorraine Li, Adhiguna Kuncoro, Aida Nematzadeh, Elena Gribovskaya, Domenic Donato, Angeliki Lazaridou, Arthur Mensch, Jean-Baptiste Lespiau, Maria Tsimpoukelli, Nikolai Grigorev, Doug Fritz, Thibault Sottiaux, Mantas Pajarskas, Toby Pohlen, Zhitao Gong, Daniel Toyama, Cyprien de~Masson~d'Autume, Yujia Li, Tayfun Terzi, Vladimir Mikulik, Igor Babuschkin, Aidan Clark, Diego de~Las~Casas, Aurelia Guy, Chris Jones,
  James Bradbury, Matthew Johnson, Blake Hechtman, Laura Weidinger, Iason Gabriel, William Isaac, Ed~Lockhart, Simon Osindero, Laura Rimell, Chris Dyer, Oriol Vinyals, Kareem Ayoub, Jeff Stanway, Lorrayne Bennett, Demis Hassabis, Koray Kavukcuoglu, and Geoffrey Irving.
\newblock Scaling language models: Methods, analysis \& insights from training gopher, 2022.

\bibitem[Ramachandran et~al.(2017)Ramachandran, Zoph, and Le]{1710.05941}
Prajit Ramachandran, Barret Zoph, and Quoc~V. Le.
\newblock Searching for activation functions, 2017.

\bibitem[Sakaguchi et~al.(2019)Sakaguchi, Bras, Bhagavatula, and Choi]{sakaguchi2019winogrande}
Keisuke Sakaguchi, Ronan~Le Bras, Chandra Bhagavatula, and Yejin Choi.
\newblock Winogrande: An adversarial winograd schema challenge at scale, 2019.

\bibitem[Sap et~al.(2019)Sap, Rashkin, Chen, LeBras, and Choi]{sap2019socialiqa}
Maarten Sap, Hannah Rashkin, Derek Chen, Ronan LeBras, and Yejin Choi.
\newblock Socialiqa: Commonsense reasoning about social interactions, 2019.

\bibitem[Scao et~al.(2022)Scao, Wang, Hesslow, Saulnier, Bekman, Bari, Biderman, Elsahar, Muennighoff, Phang, Press, Raffel, Sanh, Shen, Sutawika, Tae, Yong, Launay, and Beltagy]{2210.15424}
Teven~Le Scao, Thomas Wang, Daniel Hesslow, Lucile Saulnier, Stas Bekman, M~Saiful Bari, Stella Biderman, Hady Elsahar, Niklas Muennighoff, Jason Phang, Ofir Press, Colin Raffel, Victor Sanh, Sheng Shen, Lintang Sutawika, Jaesung Tae, Zheng~Xin Yong, Julien Launay, and Iz~Beltagy.
\newblock What language model to train if you have one million gpu hours?, 2022.

\bibitem[Shoeybi et~al.(2019)Shoeybi, Patwary, Puri, LeGresley, Casper, and Catanzaro]{shoeybi2019megatron}
Mohammad Shoeybi, Mostofa Patwary, Raul Puri, Patrick LeGresley, Jared Casper, and Bryan Catanzaro.
\newblock Megatron-lm: Training multi-billion parameter language models using model parallelism.
\newblock \emph{arXiv preprint arXiv:1909.08053}, 2019.

\bibitem[Taylor et~al.(2022)Taylor, Kardas, Cucurull, Scialom, Hartshorn, Saravia, Poulton, Kerkez, and Stojnic]{taylor2022galactica}
Ross Taylor, Marcin Kardas, Guillem Cucurull, Thomas Scialom, Anthony Hartshorn, Elvis Saravia, Andrew Poulton, Viktor Kerkez, and Robert Stojnic.
\newblock Galactica: A large language model for science, 2022.

\bibitem[Team et~al.(2023)]{mpt-7b}
MosaicML~NLP Team et~al.
\newblock Introducing mpt-7b: A new standard for open-source, commercially usable llms, 2023.
\newblock \emph{URL www. mosaicml. com/blog/mpt-7b. Accessed}, pp.\  05--05, 2023.

\bibitem[Tillet et~al.(2019)Tillet, Kung, and Cox]{Tillet2019TritonAI}
Philippe Tillet, Hsiang-Tsung Kung, and David~D. Cox.
\newblock Triton: an intermediate language and compiler for tiled neural network computations.
\newblock \emph{Proceedings of the 3rd ACM SIGPLAN International Workshop on Machine Learning and Programming Languages}, 2019.

\bibitem[Touvron et~al.(2023{\natexlab{a}})Touvron, Lavril, Izacard, Martinet, Lachaux, Lacroix, Rozi{\`e}re, Goyal, Hambro, Azhar, Rodriguez, Joulin, Grave, and Lample]{touvron2023llama}
Hugo Touvron, Thibaut Lavril, Gautier Izacard, Xavier Martinet, Marie-Anne Lachaux, Timoth{\'e}e Lacroix, Baptiste Rozi{\`e}re, Naman Goyal, Eric Hambro, Faisal Azhar, Aurelien Rodriguez, Armand Joulin, Edouard Grave, and Guillaume Lample.
\newblock Llama: Open and efficient foundation language models.
\newblock \emph{arXiv preprint arXiv:2302.13971}, 2023{\natexlab{a}}.

\bibitem[Touvron et~al.(2023{\natexlab{b}})Touvron, Martin, Stone, Albert, Almahairi, Babaei, Bashlykov, Batra, Bhargava, Bhosale, Bikel, Blecher, Ferrer, Chen, Cucurull, Esiobu, Fernandes, Fu, Fu, Fuller, Gao, Goswami, Goyal, Hartshorn, Hosseini, Hou, Inan, Kardas, Kerkez, Khabsa, Kloumann, Korenev, Koura, Lachaux, Lavril, Lee, Liskovich, Lu, Mao, Martinet, Mihaylov, Mishra, Molybog, Nie, Poulton, Reizenstein, Rungta, Saladi, Schelten, Silva, Smith, Subramanian, Tan, Tang, Taylor, Williams, Kuan, Xu, Yan, Zarov, Zhang, Fan, Kambadur, Narang, Rodriguez, Stojnic, Edunov, and Scialom]{2307.09288}
Hugo Touvron, Louis Martin, Kevin Stone, Peter Albert, Amjad Almahairi, Yasmine Babaei, Nikolay Bashlykov, Soumya Batra, Prajjwal Bhargava, Shruti Bhosale, Dan Bikel, Lukas Blecher, Cristian~Canton Ferrer, Moya Chen, Guillem Cucurull, David Esiobu, Jude Fernandes, Jeremy Fu, Wenyin Fu, Brian Fuller, Cynthia Gao, Vedanuj Goswami, Naman Goyal, Anthony Hartshorn, Saghar Hosseini, Rui Hou, Hakan Inan, Marcin Kardas, Viktor Kerkez, Madian Khabsa, Isabel Kloumann, Artem Korenev, Punit~Singh Koura, Marie-Anne Lachaux, Thibaut Lavril, Jenya Lee, Diana Liskovich, Yinghai Lu, Yuning Mao, Xavier Martinet, Todor Mihaylov, Pushkar Mishra, Igor Molybog, Yixin Nie, Andrew Poulton, Jeremy Reizenstein, Rashi Rungta, Kalyan Saladi, Alan Schelten, Ruan Silva, Eric~Michael Smith, Ranjan Subramanian, Xiaoqing~Ellen Tan, Binh Tang, Ross Taylor, Adina Williams, Jian~Xiang Kuan, Puxin Xu, Zheng Yan, Iliyan Zarov, Yuchen Zhang, Angela Fan, Melanie Kambadur, Sharan Narang, Aurelien Rodriguez, Robert Stojnic, Sergey Edunov, and Thomas
  Scialom.
\newblock Llama 2: Open foundation and fine-tuned chat models, 2023{\natexlab{b}}.

\bibitem[Vaswani et~al.(2017)Vaswani, Shazeer, Parmar, Uszkoreit, Jones, Gomez, Kaiser, and Polosukhin]{vaswani2017attention}
Ashish Vaswani, Noam Shazeer, Niki Parmar, Jakob Uszkoreit, Llion Jones, Aidan~N Gomez, {\L}ukasz Kaiser, and Illia Polosukhin.
\newblock Attention is all you need.
\newblock \emph{Advances in neural information processing systems}, 30, 2017.

\bibitem[Wang \& Komatsuzaki(2021)Wang and Komatsuzaki]{wang2021gpt}
Ben Wang and Aran Komatsuzaki.
\newblock Gpt-j-6b: A 6 billion parameter autoregressive language model, 2021.

\bibitem[Workshop et~al.(2023)Workshop, :, Scao, Fan, Akiki, Pavlick, Ilić, Hesslow, Castagné, Luccioni, Yvon, Gallé, Tow, Rush, Biderman, Webson, Ammanamanchi, Wang, Sagot, Muennighoff, del Moral, Ruwase, Bawden, Bekman, McMillan-Major, Beltagy, Nguyen, Saulnier, Tan, Suarez, Sanh, Laurençon, Jernite, Launay, Mitchell, Raffel, Gokaslan, Simhi, Soroa, Aji, Alfassy, Rogers, Nitzav, Xu, Mou, Emezue, Klamm, Leong, van Strien, Adelani, Radev, Ponferrada, Levkovizh, Kim, Natan, Toni, Dupont, Kruszewski, Pistilli, Elsahar, Benyamina, Tran, Yu, Abdulmumin, Johnson, Gonzalez-Dios, de~la Rosa, Chim, Dodge, Zhu, Chang, Frohberg, Tobing, Bhattacharjee, Almubarak, Chen, Lo, Werra, Weber, Phan, allal, Tanguy, Dey, Muñoz, Masoud, Grandury, Šaško, Huang, Coavoux, Singh, Jiang, Vu, Jauhar, Ghaleb, Subramani, Kassner, Khamis, Nguyen, Espejel, de~Gibert, Villegas, Henderson, Colombo, Amuok, Lhoest, Harliman, Bommasani, López, Ribeiro, Osei, Pyysalo, Nagel, Bose, Muhammad, Sharma, Longpre, Nikpoor, Silberberg, Pai,
  Zink, Torrent, Schick, Thrush, Danchev, Nikoulina, Laippala, Lepercq, Prabhu, Alyafeai, Talat, Raja, Heinzerling, Si, Taşar, Salesky, Mielke, Lee, Sharma, Santilli, Chaffin, Stiegler, Datta, Szczechla, Chhablani, Wang, Pandey, Strobelt, Fries, Rozen, Gao, Sutawika, Bari, Al-shaibani, Manica, Nayak, Teehan, Albanie, Shen, Ben-David, Bach, Kim, Bers, Fevry, Neeraj, Thakker, Raunak, Tang, Yong, Sun, Brody, Uri, Tojarieh, Roberts, Chung, Tae, Phang, Press, Li, Narayanan, Bourfoune, Casper, Rasley, Ryabinin, Mishra, Zhang, Shoeybi, Peyrounette, Patry, Tazi, Sanseviero, von Platen, Cornette, Lavallée, Lacroix, Rajbhandari, Gandhi, Smith, Requena, Patil, Dettmers, Baruwa, Singh, Cheveleva, Ligozat, Subramonian, Névéol, Lovering, Garrette, Tunuguntla, Reiter, Taktasheva, Voloshina, Bogdanov, Winata, Schoelkopf, Kalo, Novikova, Forde, Clive, Kasai, Kawamura, Hazan, Carpuat, Clinciu, Kim, Cheng, Serikov, Antverg, van~der Wal, Zhang, Zhang, Gehrmann, Mirkin, Pais, Shavrina, Scialom, Yun, Limisiewicz, Rieser,
  Protasov, Mikhailov, Pruksachatkun, Belinkov, Bamberger, Kasner, Rueda, Pestana, Feizpour, Khan, Faranak, Santos, Hevia, Unldreaj, Aghagol, Abdollahi, Tammour, HajiHosseini, Behroozi, Ajibade, Saxena, Ferrandis, McDuff, Contractor, Lansky, David, Kiela, Nguyen, Tan, Baylor, Ozoani, Mirza, Ononiwu, Rezanejad, Jones, Bhattacharya, Solaiman, Sedenko, Nejadgholi, Passmore, Seltzer, Sanz, Dutra, Samagaio, Elbadri, Mieskes, Gerchick, Akinlolu, McKenna, Qiu, Ghauri, Burynok, Abrar, Rajani, Elkott, Fahmy, Samuel, An, Kromann, Hao, Alizadeh, Shubber, Wang, Roy, Viguier, Le, Oyebade, Le, Yang, Nguyen, Kashyap, Palasciano, Callahan, Shukla, Miranda-Escalada, Singh, Beilharz, Wang, Brito, Zhou, Jain, Xu, Fourrier, Periñán, Molano, Yu, Manjavacas, Barth, Fuhrimann, Altay, Bayrak, Burns, Vrabec, Bello, Dash, Kang, Giorgi, Golde, Posada, Sivaraman, Bulchandani, Liu, Shinzato, de~Bykhovetz, Takeuchi, Pàmies, Castillo, Nezhurina, Sänger, Samwald, Cullan, Weinberg, Wolf, Mihaljcic, Liu, Freidank, Kang, Seelam, Dahlberg,
  Broad, Muellner, Fung, Haller, Chandrasekhar, Eisenberg, Martin, Canalli, Su, Su, Cahyawijaya, Garda, Deshmukh, Mishra, Kiblawi, Ott, Sang-aroonsiri, Kumar, Schweter, Bharati, Laud, Gigant, Kainuma, Kusa, Labrak, Bajaj, Venkatraman, Xu, Xu, Xu, Tan, Xie, Ye, Bras, Belkada, and Wolf]{workshop2023bloom}
BigScience Workshop, :, Teven~Le Scao, Angela Fan, Christopher Akiki, Ellie Pavlick, Suzana Ilić, Daniel Hesslow, Roman Castagné, Alexandra~Sasha Luccioni, François Yvon, Matthias Gallé, Jonathan Tow, Alexander~M. Rush, Stella Biderman, Albert Webson, Pawan~Sasanka Ammanamanchi, Thomas Wang, Benoît Sagot, Niklas Muennighoff, Albert~Villanova del Moral, Olatunji Ruwase, Rachel Bawden, Stas Bekman, Angelina McMillan-Major, Iz~Beltagy, Huu Nguyen, Lucile Saulnier, Samson Tan, Pedro~Ortiz Suarez, Victor Sanh, Hugo Laurençon, Yacine Jernite, Julien Launay, Margaret Mitchell, Colin Raffel, Aaron Gokaslan, Adi Simhi, Aitor Soroa, Alham~Fikri Aji, Amit Alfassy, Anna Rogers, Ariel~Kreisberg Nitzav, Canwen Xu, Chenghao Mou, Chris Emezue, Christopher Klamm, Colin Leong, Daniel van Strien, David~Ifeoluwa Adelani, Dragomir Radev, Eduardo~González Ponferrada, Efrat Levkovizh, Ethan Kim, Eyal~Bar Natan, Francesco~De Toni, Gérard Dupont, Germán Kruszewski, Giada Pistilli, Hady Elsahar, Hamza Benyamina, Hieu Tran,
  Ian Yu, Idris Abdulmumin, Isaac Johnson, Itziar Gonzalez-Dios, Javier de~la Rosa, Jenny Chim, Jesse Dodge, Jian Zhu, Jonathan Chang, Jörg Frohberg, Joseph Tobing, Joydeep Bhattacharjee, Khalid Almubarak, Kimbo Chen, Kyle Lo, Leandro~Von Werra, Leon Weber, Long Phan, Loubna~Ben allal, Ludovic Tanguy, Manan Dey, Manuel~Romero Muñoz, Maraim Masoud, María Grandury, Mario Šaško, Max Huang, Maximin Coavoux, Mayank Singh, Mike Tian-Jian Jiang, Minh~Chien Vu, Mohammad~A. Jauhar, Mustafa Ghaleb, Nishant Subramani, Nora Kassner, Nurulaqilla Khamis, Olivier Nguyen, Omar Espejel, Ona de~Gibert, Paulo Villegas, Peter Henderson, Pierre Colombo, Priscilla Amuok, Quentin Lhoest, Rheza Harliman, Rishi Bommasani, Roberto~Luis López, Rui Ribeiro, Salomey Osei, Sampo Pyysalo, Sebastian Nagel, Shamik Bose, Shamsuddeen~Hassan Muhammad, Shanya Sharma, Shayne Longpre, Somaieh Nikpoor, Stanislav Silberberg, Suhas Pai, Sydney Zink, Tiago~Timponi Torrent, Timo Schick, Tristan Thrush, Valentin Danchev, Vassilina Nikoulina,
  Veronika Laippala, Violette Lepercq, Vrinda Prabhu, Zaid Alyafeai, Zeerak Talat, Arun Raja, Benjamin Heinzerling, Chenglei Si, Davut~Emre Taşar, Elizabeth Salesky, Sabrina~J. Mielke, Wilson~Y. Lee, Abheesht Sharma, Andrea Santilli, Antoine Chaffin, Arnaud Stiegler, Debajyoti Datta, Eliza Szczechla, Gunjan Chhablani, Han Wang, Harshit Pandey, Hendrik Strobelt, Jason~Alan Fries, Jos Rozen, Leo Gao, Lintang Sutawika, M~Saiful Bari, Maged~S. Al-shaibani, Matteo Manica, Nihal Nayak, Ryan Teehan, Samuel Albanie, Sheng Shen, Srulik Ben-David, Stephen~H. Bach, Taewoon Kim, Tali Bers, Thibault Fevry, Trishala Neeraj, Urmish Thakker, Vikas Raunak, Xiangru Tang, Zheng-Xin Yong, Zhiqing Sun, Shaked Brody, Yallow Uri, Hadar Tojarieh, Adam Roberts, Hyung~Won Chung, Jaesung Tae, Jason Phang, Ofir Press, Conglong Li, Deepak Narayanan, Hatim Bourfoune, Jared Casper, Jeff Rasley, Max Ryabinin, Mayank Mishra, Minjia Zhang, Mohammad Shoeybi, Myriam Peyrounette, Nicolas Patry, Nouamane Tazi, Omar Sanseviero, Patrick von
  Platen, Pierre Cornette, Pierre~François Lavallée, Rémi Lacroix, Samyam Rajbhandari, Sanchit Gandhi, Shaden Smith, Stéphane Requena, Suraj Patil, Tim Dettmers, Ahmed Baruwa, Amanpreet Singh, Anastasia Cheveleva, Anne-Laure Ligozat, Arjun Subramonian, Aurélie Névéol, Charles Lovering, Dan Garrette, Deepak Tunuguntla, Ehud Reiter, Ekaterina Taktasheva, Ekaterina Voloshina, Eli Bogdanov, Genta~Indra Winata, Hailey Schoelkopf, Jan-Christoph Kalo, Jekaterina Novikova, Jessica~Zosa Forde, Jordan Clive, Jungo Kasai, Ken Kawamura, Liam Hazan, Marine Carpuat, Miruna Clinciu, Najoung Kim, Newton Cheng, Oleg Serikov, Omer Antverg, Oskar van~der Wal, Rui Zhang, Ruochen Zhang, Sebastian Gehrmann, Shachar Mirkin, Shani Pais, Tatiana Shavrina, Thomas Scialom, Tian Yun, Tomasz Limisiewicz, Verena Rieser, Vitaly Protasov, Vladislav Mikhailov, Yada Pruksachatkun, Yonatan Belinkov, Zachary Bamberger, Zdeněk Kasner, Alice Rueda, Amanda Pestana, Amir Feizpour, Ammar Khan, Amy Faranak, Ana Santos, Anthony Hevia, Antigona
  Unldreaj, Arash Aghagol, Arezoo Abdollahi, Aycha Tammour, Azadeh HajiHosseini, Bahareh Behroozi, Benjamin Ajibade, Bharat Saxena, Carlos~Muñoz Ferrandis, Daniel McDuff, Danish Contractor, David Lansky, Davis David, Douwe Kiela, Duong~A. Nguyen, Edward Tan, Emi Baylor, Ezinwanne Ozoani, Fatima Mirza, Frankline Ononiwu, Habib Rezanejad, Hessie Jones, Indrani Bhattacharya, Irene Solaiman, Irina Sedenko, Isar Nejadgholi, Jesse Passmore, Josh Seltzer, Julio~Bonis Sanz, Livia Dutra, Mairon Samagaio, Maraim Elbadri, Margot Mieskes, Marissa Gerchick, Martha Akinlolu, Michael McKenna, Mike Qiu, Muhammed Ghauri, Mykola Burynok, Nafis Abrar, Nazneen Rajani, Nour Elkott, Nour Fahmy, Olanrewaju Samuel, Ran An, Rasmus Kromann, Ryan Hao, Samira Alizadeh, Sarmad Shubber, Silas Wang, Sourav Roy, Sylvain Viguier, Thanh Le, Tobi Oyebade, Trieu Le, Yoyo Yang, Zach Nguyen, Abhinav~Ramesh Kashyap, Alfredo Palasciano, Alison Callahan, Anima Shukla, Antonio Miranda-Escalada, Ayush Singh, Benjamin Beilharz, Bo~Wang, Caio Brito,
  Chenxi Zhou, Chirag Jain, Chuxin Xu, Clémentine Fourrier, Daniel~León Periñán, Daniel Molano, Dian Yu, Enrique Manjavacas, Fabio Barth, Florian Fuhrimann, Gabriel Altay, Giyaseddin Bayrak, Gully Burns, Helena~U. Vrabec, Imane Bello, Ishani Dash, Jihyun Kang, John Giorgi, Jonas Golde, Jose~David Posada, Karthik~Rangasai Sivaraman, Lokesh Bulchandani, Lu~Liu, Luisa Shinzato, Madeleine~Hahn de~Bykhovetz, Maiko Takeuchi, Marc Pàmies, Maria~A Castillo, Marianna Nezhurina, Mario Sänger, Matthias Samwald, Michael Cullan, Michael Weinberg, Michiel~De Wolf, Mina Mihaljcic, Minna Liu, Moritz Freidank, Myungsun Kang, Natasha Seelam, Nathan Dahlberg, Nicholas~Michio Broad, Nikolaus Muellner, Pascale Fung, Patrick Haller, Ramya Chandrasekhar, Renata Eisenberg, Robert Martin, Rodrigo Canalli, Rosaline Su, Ruisi Su, Samuel Cahyawijaya, Samuele Garda, Shlok~S Deshmukh, Shubhanshu Mishra, Sid Kiblawi, Simon Ott, Sinee Sang-aroonsiri, Srishti Kumar, Stefan Schweter, Sushil Bharati, Tanmay Laud, Théo Gigant, Tomoya
  Kainuma, Wojciech Kusa, Yanis Labrak, Yash~Shailesh Bajaj, Yash Venkatraman, Yifan Xu, Yingxin Xu, Yu~Xu, Zhe Tan, Zhongli Xie, Zifan Ye, Mathilde Bras, Younes Belkada, and Thomas Wolf.
\newblock Bloom: A 176b-parameter open-access multilingual language model, 2023.

\bibitem[Zellers et~al.(2019)Zellers, Holtzman, Bisk, Farhadi, and Choi]{zellers2019hellaswag}
Rowan Zellers, Ari Holtzman, Yonatan Bisk, Ali Farhadi, and Yejin Choi.
\newblock Hellaswag: Can a machine really finish your sentence?, 2019.

\bibitem[Zeng et~al.(2022)Zeng, Liu, Du, Wang, Lai, Ding, Yang, Xu, Zheng, Xia, et~al.]{zeng2022glm}
Aohan Zeng, Xiao Liu, Zhengxiao Du, Zihan Wang, Hanyu Lai, Ming Ding, Zhuoyi Yang, Yifan Xu, Wendi Zheng, Xiao Xia, et~al.
\newblock Glm-130b: An open bilingual pre-trained model.
\newblock \emph{arXiv preprint arXiv:2210.02414}, 2022.

\bibitem[Zhang et~al.(2022)Zhang, Roller, Goyal, Artetxe, Chen, Chen, Dewan, Diab, Li, Lin, Mihaylov, Ott, Shleifer, Shuster, Simig, Koura, Sridhar, Wang, and Zettlemoyer]{zhang2022opt}
Susan Zhang, Stephen Roller, Naman Goyal, Mikel Artetxe, Moya Chen, Shuohui Chen, Christopher Dewan, Mona Diab, Xian Li, Xi~Victoria Lin, Todor Mihaylov, Myle Ott, Sam Shleifer, Kurt Shuster, Daniel Simig, Punit~Singh Koura, Anjali Sridhar, Tianlu Wang, and Luke Zettlemoyer.
\newblock Opt: Open pre-trained transformer language models, 2022.

\bibitem[Zhao et~al.(2023)Zhao, Gu, Varma, Luo, Huang, Xu, Wright, Shojanazeri, Ott, Shleifer, et~al.]{zhao2023pytorch}
Yanli Zhao, Andrew Gu, Rohan Varma, Liang Luo, Chien-Chin Huang, Min Xu, Less Wright, Hamid Shojanazeri, Myle Ott, Sam Shleifer, et~al.
\newblock Pytorch fsdp: experiences on scaling fully sharded data parallel.
\newblock \emph{arXiv preprint arXiv:2304.11277}, 2023.

\bibitem[Zheng et~al.(2022)Zheng, Wang, and Kong]{zheng2022linear}
Lin Zheng, Chong Wang, and Lingpeng Kong.
\newblock Linear complexity randomized self-attention mechanism.
\newblock In \emph{International Conference on Machine Learning}, pp.\  27011--27041. PMLR, 2022.

\bibitem[Zheng et~al.(2023)Zheng, Yuan, Wang, and Kong]{zheng2023efficient}
Lin Zheng, Jianbo Yuan, Chong Wang, and Lingpeng Kong.
\newblock Efficient attention via control variates.
\newblock In \emph{International Conference on Learning Representations}, 2023.
\newblock URL \url{https://openreview.net/forum?id=G-uNfHKrj46}.

\end{thebibliography}
\bibliographystyle{iclr2024_conference}

\appendix
\newpage
\begin{center}
\textbf{\large Appendix}
\end{center}

\section{Model}
\label{app:model}
We present distinct model variants of the TransNormerLLM architecture, delineating their respective configurations with regard to parameters, layers, attention heads, and hidden dimensions. The detailed specifications are meticulously tabulated in Table~\ref{transnormer_models}.

\begin{table}[htbp]
\small
    \caption{\textbf{TransNormerLLM Model Variants.} }
    \label{transnormer_models}
    \centering
    \renewcommand{\arraystretch}{1}
    \setlength{\tabcolsep}{0.32cm}
    \begin{tabular}{cccccc}
    \hline
    
        Model Size & Non-Embedding Params & Layers & Hidden Dim & Heads & Equivalent Models \\ \hline
        385M & 384,974,848 & 24 & 1024 & 8 & Pythia-410M  \\ 
        1B & 992,165,888 & 16 & 2048 & 16 & Pythia-1B  \\ 
        3B & 2,876,006,400 & 32 & 2560 & 20 & Pythia-2.8B  \\ 
        7B & 6,780,547,072 & 30 & 4096 & 32 & LLAMA-6.7B  \\ 
        13B & 12,620,195,840  & 36 & 5120 & 40 & LLAMA-13B  \\ 
        65B & 63,528,009,728 & 72 & 8192 & 64 & LLAMA-65B  \\ 
        175B & 173,356,498,944 & 88 & 12288 & 96 & GPT-3  \\ 
        \hline
    \end{tabular}
\end{table}

\section{Lightning Attention}
\label{app:lightning}
We present the algorithm details of Lightning Attention includes forward pass and backward pass in Algorithm \ref{algo:lightning attention fw pseudo} and \ref{algo:lightning attention bw pseudo}, respectively.

\begin{algorithm}
\small
    \caption{Lightning Attention Forward Pass}
    \label{algo:lightning attention fw pseudo}
    \begin{algorithmic}
    \State{\textbf{Input:} $\mathbf Q,\mathbf K,\mathbf V \in \mathbb{R}^{n \times d}$, attention mask $\mathbf{M }\in \mathbb{R}^{n \times n} $, 
    block sizes $B_c,B_r$;}
    \State{\textbf{Initialize:} $\mathbf O=\mathbf 0 \in \mathbb{R}^{n \times d}$;}
     \State{Divide $\mathbf Q$ into $T_r = \frac{n}{B_r}$ blocks $\mathbf Q_1, \mathbf Q_2, ...\mathbf Q_{T_r}$ of size $B_r \times d$ each. }
     \State{Divide $\mathbf K,\mathbf V$ into $T_c = \frac{n}{B_c}$ blocks $\mathbf K_1, \mathbf K_2, ...\mathbf K_{T_c}, \mathbf V_1, \mathbf V_2, ...\mathbf V_{T_c} $ of size $B_c \times d$ each.}
     \State{Divide $\mathbf O$ into $T_r = \frac{n}{B_r}$ blocks $\mathbf O_1, \mathbf O_2, ...\mathbf O_{T_r}$ of size $B_r \times d$ each.}
     \State{Divide $\mathbf M$ into $T_r \times T_c$ blocks $\mathbf M_{11}, \mathbf M_{12}, ...\mathbf M_{T_r,T_c}$ of size $B_r \times B_c$ each.}
    \For{$1 \leq i \leq T_r$}
        \State{Load $\mathbf Q_i \in \mathbb{R}^{B_r \times d}$ from HBM to on-chip SRAM.}
        \State{Initialize $ \mathbf{O}_i= \mathbf 0 \in \mathbb{R}^{B_r \times d}$ on SRAM.}
        \For{$1 \leq j \leq T_c$}
            \State{Load $\mathbf K_j, \mathbf V_j \in \mathbb{R}^{B_c \times d}$ from HBM to on-chip SRAM.}
            \State{Load $\mathbf M_{ij} \in \mathbb{R}^{B_c \times B_c}$ from HBM to on-chip SRAM.}
            \State{On chip, compute $\mathbf A_{ij} = [\mathbf Q_i \mathbf K_j^{\top }] \odot \mathbf M_{ij}\in \mathbb{R}^{B_r \times B_c}$.}
            \State{On chip, compute $\mathbf{O}_i = \mathbf{O}_i + \mathbf A_{ij}\mathbf V_j \in \mathbb{R}^{B_r \times d}$.}
      \EndFor
      \State{Write $\mathbf O_i$ to HBM as the $i$-th block of $\mathbf O$.}
      \EndFor
      \State{return $\mathbf O$}
\end{algorithmic}
\end{algorithm}

\begin{algorithm}
\small
    \caption{Lightning Attention Backward Pass}
    \label{algo:lightning attention bw pseudo}
    \begin{algorithmic}
    \State{\textbf{Input:} $\mathbf Q,\mathbf K,\mathbf V,\mathbf{dO} \in \mathbb{R}^{n \times d}$, attention mask $\mathbf{M }\in \mathbb{R}^{n \times n} $,  on-chip SRAM of size $M$, block sizes $B_c,B_r$;}
    \State{\textbf{Initialize:} $\mathbf{dQ}=\mathbf{dK}=\mathbf{dV}=\mathbf 0 \in \mathbb{R}^{n \times d}$;}
     \State{Divide $\mathbf Q$ into $T_r = \frac{n}{B_r}$ blocks $\mathbf Q_1, \mathbf Q_2, ...\mathbf Q_{T_r}$ of size $B_r \times d$ each. }
     \State{Divide $\mathbf K,\mathbf V$ into $T_c = \frac{n}{B_c}$ blocks $\mathbf K_1, \mathbf K_2, ...\mathbf K_{T_c}, \mathbf V_1, \mathbf V_2, ...\mathbf V_{T_c} $ of size $B_c \times d$ each.}
     \State{Divide $\mathbf O,\mathbf {dO}$ into $T_r = \frac{n}{B_r}$ blocks $\mathbf O_1, \mathbf O_2, ...\mathbf O_{T_r},\mathbf {dO_1}, \mathbf {dO_2}, ...\mathbf {dO_{T_r}}$ of size $B_r \times d$ each}
     \State{Divide $\mathbf M$ into $T_r \times T_c$ blocks $\mathbf M_{11}, \mathbf M_{12}, ...\mathbf M_{T_r,T_c}$ of size $B_r \times B_c$ each.}
    \For{$1 \leq j \leq T_c$}
        \State{Load $\mathbf K_j, \mathbf V_j \in \mathbb{R}^{B_c \times d}$ from HBM to on-chip SRAM.}
        \State{Initialize $ \mathbf{dK}_j=\mathbf{dV}_j= \mathbf 0 \in \mathbb{R}^{B_c \times d}$ on SRAM.}
        \For{$1 \leq i \leq T_r$}
             \State{Load $\mathbf Q_i,\mathbf O_i,\mathbf {dO}_i \in \mathbb{R}^{B_r \times d}$ from HBM to on-chip SRAM.}
             \State{Load $\mathbf M_{ij} \in \mathbb{R}^{B_c \times B_c}$ from HBM to on-chip SRAM.}
             \State{Initialize $ \mathbf{dK}_j=\mathbf{dV}_j= \mathbf 0 \in \mathbb{R}^{B_c \times d}$ on SRAM.}           
            \State{On chip, compute $\mathbf A_{ij} = [\mathbf Q_i \mathbf K_j^{\top}]\odot \mathbf M_{ij} \in \mathbb{R}^{B_r \times B_c}$.}
              \State{On chip, compute $\mathbf {dV}_{j} =  \mathbf {dV}_{j}+\mathbf A_{ij}^{\top} \mathbf{dO}_i \in \mathbb{R}^{B_c \times d}$.}
            \State{On chip, compute $\mathbf {dA}_{ij} =  [\mathbf {dO}_{i}\mathbf V_j^{\top}] \odot \mathbf M_{ij} \in \mathbb{R}^{B_r \times B_c}$.}
           \State{On chip, compute $\mathbf {dK}_{j} =  \mathbf {dk}_{j}+\mathbf {dA}_{ij}^{\top} \mathbf{V}_j \in \mathbb{R}^{B_c \times d}$.}
             \State{Load $\mathbf{dQ}_i$ from HBM to SRAM, then on chip, compute $\mathbf {dQ}_{i} =  \mathbf {dK}_{i}+\mathbf {dA}_{ij} \mathbf{K}_j \in \mathbb{R}^{B_r \times d}$,}
             \State{write back to HBM.}
      \EndFor
     \State{Write $\mathbf {dK}_j,\mathbf {dV}_j$ to HBM as the $j$-th block of $\mathbf {dK}, \mathbf {dV}$.}
      \EndFor
      \State{retun $\mathbf {dQ, dK, dV}$}
\end{algorithmic}
\end{algorithm}

\section{Proving robust inference algorithm}
\label{app:robustinfer}
We will use induction to prove: $[\mathbf {kv}]_t=\lambda^{-t}[{\mathbf {\overline{kv}}}]_t$.

\small
    \textbf{Base Case ($n=1$):}
    \begin{equation}
    \begin{aligned}
    \relax[\mathbf{kv}]_1 &=([\mathbf {kv}]_0+\mathbf {k_1} \lambda^{-1} \mathbf {v}_1^{\top})\\
    &=\lambda^{-1}(\mathbf {k_1}  \mathbf {v}_1^{\top})\\
    &= \lambda^{-1}[{\mathbf {\overline{kv}}}]_1 .
    \end{aligned}
    \end{equation}
Assume the statement holds for $n=m-1$, i.e., $[\mathbf {kv}]_{m-1}=\lambda^{-(m-1)}[{\mathbf {\overline{kv}}}]_{m-1}$. Then, when $n=m$:
    \begin{equation}
    \begin{aligned}
    \relax[\mathbf{kv}]_m &= [\mathbf {kv}]_{m-1} + \mathbf {k_m} \lambda^{-m} \mathbf {v}_m^{\top}\\
    &=\lambda^{-(m-1)}[{\mathbf {\overline{kv}}}]_{m-1} + \mathbf {k_m} \lambda^{-m} \mathbf {v}_m^{\top}\\
    &=\lambda^{-m}(\lambda [{\mathbf {\overline{kv}}}]_{m-1} + \mathbf {k_m}\mathbf {v}_m^{\top})\\
    &=\lambda^{-m} [{\mathbf {\overline{kv}}}]_m,
    \end{aligned}
    \end{equation}
\normalsize
the statement holds. Therefore, by induction, the statement holds for all $n\geq 1$.

Thus, both the Origin Inference Algorithm and the Robust Inference Algorithm yield the same results.

\section{Corpus}
\label{app:corpus}
\begin{figure}[htbp]
    \centering
    \vspace{-0mm}
        \includegraphics[width=\textwidth]{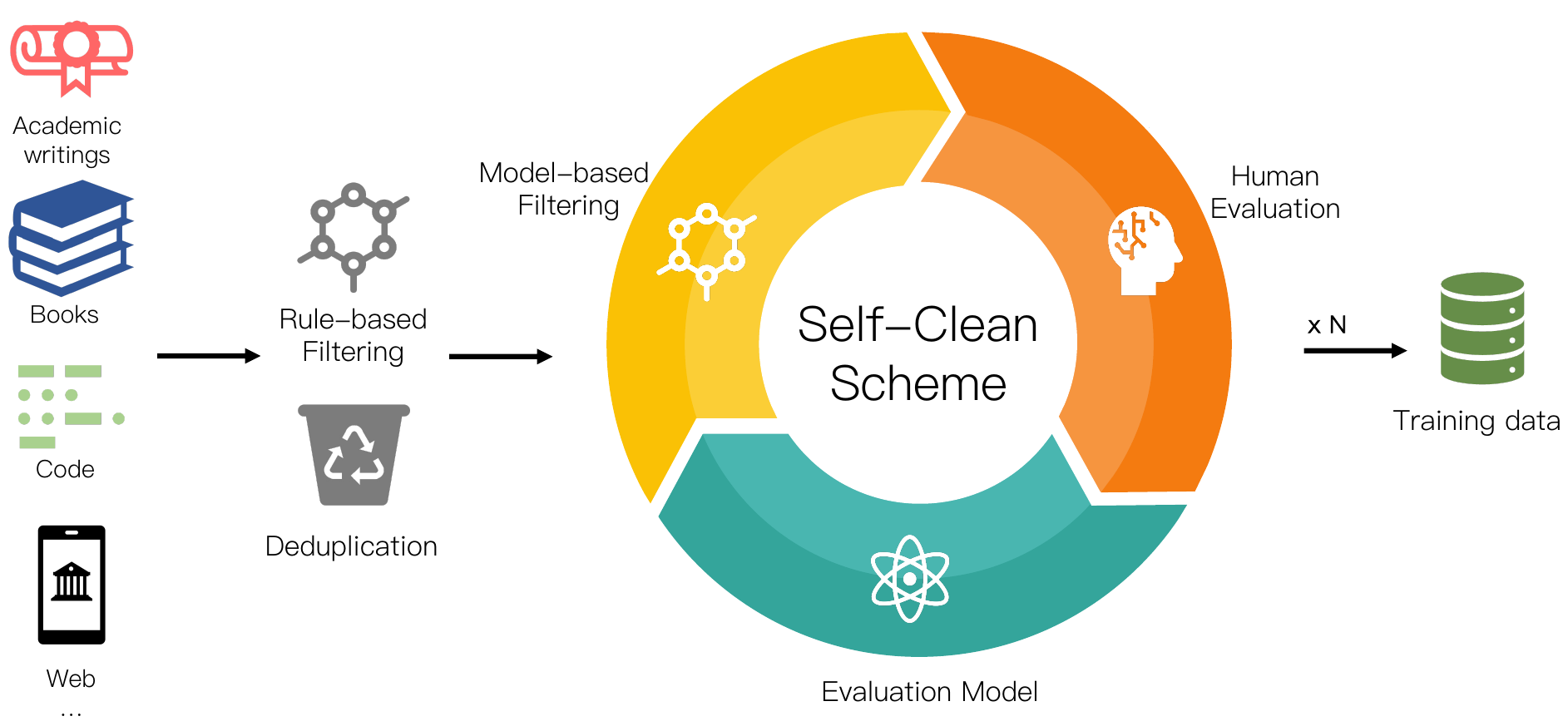}
        \vspace{-4mm}
        \caption{\textbf{Data Preprocess Procedure.} The collected data undergoes a process of rule-based filtering and deduplication, followed by our self-clean data processing strategy: model-based filtering, human evaluation, and evaluation model. After several iterations of the above cycle, we obtain high-quality training data at around 2T tokens.}
        \label{fig:data_preprocess}
\end{figure}

We gather an extensive corpus of publicly accessible text from the internet, totaling over $700$TB in size. The collected data are processed by our data preprocessing procedure as shown in Figure~\ref{fig:data_preprocess}, leaving a $6$TB cleaned corpus with roughly 2 trillion tokens. We categorize our data sources to provide better transparency and understanding. The specifics of these categories are outlined in Table~\ref{tab:pretraing_data}. 

\subsection{Data Preprocessing}

Our data preprocessing procedure consists of three steps: 1). rule-based filtering, 2). deduplication, and 3). a self-cleaning scheme. Before being added to the training corpus, the cleaned corpus needs to be evaluated by humans.

\paragraph{Rule-based filtering}
The rules we used to filter our collected data are listed as follows:
\begin{itemize}[noitemsep,topsep=0pt,parsep=0pt,partopsep=0pt]
    \item \emph{Removal of HTML Tags and URLs:} The initial step in our process is the elimination of HTML tags and web URLs from the text. This is achieved through regular expression techniques that identify these patterns and remove them, ensuring the language model focuses on meaningful textual content.
    \item \emph{Elimination of Useless or Abnormal Strings:} Subsequently, the cleaned dataset undergoes a second layer of refinement where strings that do not provide value, such as aberrant strings or garbled text, are identified and excised. This process relies on predefined rules that categorize certain string patterns as non-contributing elements.
    \item \emph{Deduplication of Punctuation Marks:} We address the problem of redundant punctuation marks in the data. Multiple consecutive punctuation marks can distort the natural flow and structure of sentences when training the model. We employ a rule-based system that trims these duplications down to a single instance of each punctuation mark.
    \item \emph{Handling Special Characters:} Unusual or special characters that are not commonly part of the language's text corpus are identified and either removed or replaced with a standardized representation.
    \item \emph{Number Standardization:} Numerical figures may be presented in various formats across different texts. These numbers are standardized into a common format to maintain consistency.
    \item \emph{Preservation of Markdown/LaTeX Formats:} While removing non-textual elements, exceptions are made for texts in Markdown and LaTeX formats. Given their structured nature and ubiquitous use in academia and documentation, preserving these formats can enhance the model's ability to understand and generate similarly formatted text.
\end{itemize}

\paragraph{Deduplication}
To ensure the uniqueness of our data and avert the risk of overfitting, we employ an efficient de-duplication strategy at the document or line level using MinHash and Locality-Sensitive Hashing (LSH) algorithms. This combination of MinHash and LSH ensures a balance between computational efficiency and accuracy in the deduplication process, providing a robust mechanism for data deduplication and text watermark removal. 

\paragraph{Self-cleaning scheme}
Our data self-cleaning process involves an iterative loop of the following three steps to continuously refine and enhance the quality of our dataset. An issue of using model-based data filters is that the filtered data will have a similar distribution as the evaluation model, which may have a significant impact on the diversity of the training data. Assuming that the majority of the pre-processed data is of high quality, we can train an evaluation model on the entire set of pre-processed data, and the model will automatically smooth the data manifold distribution and outlet low-quality data while retaining the majority of the diversities. 

The self-cleaning scheme unfolds as follows:
\begin{itemize}[noitemsep,topsep=0pt,parsep=0pt,partopsep=0pt]
    \item \emph{Evaluation Model:} We train a 385M model on the pre-processed corpus to act as a data quality filter. 
    \item \emph{Model-Based Data Filtering:} We use the evaluation model to assess each piece of data with perplexity. Only data achieving a score above a certain threshold is preserved for the next step. Low-quality data are weeded out at this stage.
    \item \emph{Human Evaluation:} We sample a small portion of the filtered data and manually evaluate the quality. 
\end{itemize}

These steps are repeated in cycles, with each iteration improving the overall quality of the data and ensuring the resulting model is trained on relevant, high-quality text. This self-cleaning process provides a robust mechanism for maintaining data integrity, thereby enhancing the performance of the resulting language model.

\begin{table}[t]
\small
    \caption{\textbf{Statistics of our corpus.} For each category, we list the number of epochs performed on the subset when training on the 2 trillion tokens, as well as the number of tokens and disk sizes. We also list the table on the right according to the language distribution.  }
    \label{tab:pretraing_data}
    \centering
    \begin{minipage}{0.6\textwidth}
        \centering
        \setlength{\tabcolsep}{3.2mm}
        \renewcommand{\arraystretch}{1.0}
        \begin{tabular}{lcrr}
        \hline
        Dataset & Epochs & Tokens & Disk size \\ \hline
        Academic Writings & 1.53 & 200 B & 672 GB \\ 
        Books & 2.49 & 198 B & 723 GB \\
        Code & 0.44 & 689 B & 1.4 TB \\
        Encyclopedia & 1.51 & 5 B & 18 GB \\
        Filtered Webpages & 1.00 & 882 B & 3.1 TB \\
        Others & 0.63 & 52 B & 154 GB \\ \hline
        Total & - & 2026 B  & 6 TB \\ \hline
        \end{tabular}
    \end{minipage}%
    \hfill
    \begin{minipage}{0.4\textwidth}
        \centering
        \setlength{\tabcolsep}{3.4mm}
        \renewcommand{\arraystretch}{1.27}
        \begin{tabular}{lrr}
        \hline
        \\[-1em]
        Language & Tokens & Disk size \\ \hline
        English & 743 B & 2.9 TB \\
        Chiese & 555 B & 1.7 TB \\
        Code & 689 B & 1.4 TB \\
        Others & 39 B & 89 GB \\ \hline
        Total & 2026 B & 6 TB \\ \hline
        \end{tabular}
    \end{minipage}%
\end{table}


\subsection{Tokenization}
We tokenize the data with the Byte-Pair Encoding (BPE) algorithm. Notably, to enhance compatibility with Chinese language content, a significant number of common and uncommon Chinese characters have been incorporated into our vocabulary. In cases where vocabulary items are not present in the dictionary, the words are broken down into their constituent UTF-8 characters. This strategy ensures comprehensive coverage and flexibility for diverse linguistic input during model training.

\section{Additional Experimental Results}

\subsection{Model Parallelism on TransNormerLLM}
\label{app:model_para}
We conduct a series of experiments with a 7B TransNormerLLM model to investigate the performance of model parallelism on TransNormerLLM in terms of speed and memory. These tests are carried out on a single Nvidia DGX node that houses eight A100 80G GPUs linked by NVLink. In this experiment, FSDP is enabled and Flash Attention~\citep{dao2022flashattention} is used on the Transformer. Table~\ref{tab:model_parallel} shows the results for training speed and memory consumption.

It can be seen that model parallelism has a significant effect on memory conservation, as increasing the number of partitions for the model results in lower memory consumption per GPU. Due to NVLink constraints, we kept the dimension of model parallelism within 8 in all of our experiments. The TransNormerLLM-7B model requires only 24.1GB of memory on a single GPU when the model parallel size is set to 8, representing a significant memory reduction of 62.3\% when compared to the model parallel size of 1. In comparison, the Transformer-7B model consumes 28.7GB of memory under the same configuration. While model parallelism conserves memory, it is worth noting that training speed is only marginally reduced. TransNormerLLM consistently outperforms Transformer by a wide margin.

\begin{table}[htbp]
\centering
\small
\setlength{\tabcolsep}{2.4mm}
\caption{\textbf{Model Parallelism Performance.} We compare the model parallelism performance of Transformer-7B with Flash Attention and TransNormerLLM-7B with Lightning Attention on a single A100 node with NVLink. All experiments use a batch size of 2 and a context length of 2048.}
\label{tab:model_parallel}
\begin{tabular}{cccccc}
\toprule
\small
Model & Model Parallel Size & Tokens/s & Allocated Memory/GPU & Memory Saved \\
\midrule
\multirow{4}{*}{Transformer-7B} & 1 & 26896.1 & 66.3 GB &  - \\
 & 2 & 24973.7 & 44.6 GB & 32.7\% \\
 & 4 & 22375.8 & 40.2 GB & 39.4\% \\
 & 8 & 19973.6 & 28.7 GB & 56.7\% \\
\midrule
\multirow{4}{*}{TransNormerLLM-7B} & 1 & 32048.6 & 64.0 GB & - \\
 & 2 & 29750.4 & 41.0 GB & 35.9\% \\
 & 4 & 27885.2 & 36.3 GB & 43.3\% \\
 & 8 & 24280.0 & 24.1 GB & 62.3\% \\
\bottomrule
\end{tabular}
\end{table}

\subsection{Stress Tests on Model Size and Context Length}
\label{app:size_length}
A series of stress tests are performed to assess the efficacy of the designed system optimization strategy. The model is scaled up to 175B, which is the largest released version of the TransNormerLLM model. However, this augmentation poses significant training challenges. We use a wide range of distributed training techniques to effectively train such a large model, with the goal of reducing GPU memory consumption while increasing computational and communication efficiencies. To ensure the feasibility of training these massive TransNormerLLM models, Lightning Attention, FSDP, Model Parallelism, AMP, and Activation Checkpointing are used. For the Transformer models, we use Flash Attention~\citep{dao2022flashattention} in all experiments.

\paragraph{Model Size}

We perform training experiments on variously sized Transformer and TransNormerLLM models using a large-scale A100 80G GPU cluster, as shown in Table~\ref{tab:175b_speed}. To achieve the maximum speed for various model sizes, we keep the context length constant at 2048 and increased the batch size until we reached the GPU memory limit. TransNormerLLMs consistently outperform their Transformer counterparts in terms of computation speed. This observation validates the TransNormerLLM model's advantageous linear computational complexity, reinforcing its efficacy.

\begin{table}[htbp]
\small
\centering
\setlength{\tabcolsep}{6mm}
\caption{\textbf{Efficiency of training models with different sizes.} For comparative purposes, we keep the context length fixed at 2048 and increased the batch size for both transformer and TransNormerLLM to achieve their maximum speeds without encountering out-of-memory issues.}
\label{tab:175b_speed}
\begin{tabular}{cccc}
\toprule
Model & Model Size & Tokens/sec/GPU & Allocated Memory/GPU \\
\midrule
\multirow{4}{*}{Transformer} & 7B & 3362.7 & 72.5 GB \\
 & 13B & 1735.6 & 70.6 GB \\
 & 65B & 318.2 & 73.2 GB \\
 & 175B & 106.2 & 69.5 GB \\
\midrule
\multirow{4}{*}{TransNormerLLM} & 7B & 4081.0 & 71.9 GB \\
 & 13B & 2104.3 & 73.8 GB \\
 & 65B & 406.9 & 69.4 GB \\
 & 175B & 136.6 & 70.3 GB \\
\bottomrule
\end{tabular}
\end{table}

\paragraph{Context Length}
One of the strengths of TransNormerLLM lies in its utilization of linear attention computation, which exhibits computational and storage complexities linearly correlated with the sequence length. 
To validate this outstanding characteristic of TransNormerLLM, we conduct training experiments on Transformer and TransNormerLLM models with varying parameter sizes. While maintaining a batch size of 1, we aim to maximize the context length. All experiments run on a small cluster with 64 A100 GPUs. The results, as presented in Table \ref{tab:context_length}, demonstrate the remarkable long context length training capability of TransNormerLLM. Under comparable computational resources, the TransNormerLLM model exhibits the ability to train with longer context lengths compared to conventional Transformer models and achieve higher computational speeds in the process.

\begin{table}[htbp]
\small
\centering
\setlength{\tabcolsep}{3mm}
\caption{\textbf{Maximum context length for training Transformer and TransNormerLLM.} We compare the maximum context lengths with different model sizes between Transformer and TransNormerLLM on 64 A100 80G GPUs. All experiments use a batch size of 1.}
\label{tab:context_length}
\begin{tabular}{ccccc}
\toprule
Model & Model Size & Context Length & Relative Speed & Allocated Memory/GPU \\
\midrule
\multirow{4}{*}{Transformer} & 7B & 37K & 1 & 71.1 GB \\
 & 13B & 24K & 1 & 68.0 GB \\
 & 65B & 19K & 1 & 73.3 GB \\
 & 175B & 10K & 1  & 66.9 GB \\
\midrule
\multirow{4}{*}{TransNormerLLM} & 7B & 48K & 1.21 & 65.8 GB \\
 & 13B & 35K & 1.23 & 61.0 GB \\
 & 65B & 23K & 1.29 & 68.2 GB \\
 & 175B & 12K & 1.35 & 63.5 GB \\
\bottomrule
\end{tabular}
\end{table}

\end{document}